\documentclass[sigconf]{acmart}
\AtBeginDocument{%
  }

\usepackage{enumitem}
\usepackage{placeins}

\setcopyright{acmlicensed}
\copyrightyear{2018}
\acmYear{2018}
\acmDOI{XXXXXXX.XXXXXXX}

\newcommand{\calD}{\mathcal{D}}

\newcommand{\calP}{\mathcal{P}}
\newcommand{\calX}{\mathcal{X}}
\newcommand{\calL}{\mathcal{L}}

\newcommand{\bA}{\mathbf{A}}
\newcommand{\bb}{\mathbf{b}}
\newcommand{\bD}{\mathbf{D}}

\newcommand{\bH}{\mathbf{H}}
\newcommand{\by}{\mathbf{y}}
\newcommand{\bC}{\mathbf{C}}
\newcommand{\bW}{\mathbf{W}}
\newcommand{\bB}{\mathbf{B}}
\newcommand{\ba}{\mathbf{a}}

\newcommand{\bK}{\mathbf{K}}
\newcommand{\bx}{\mathbf{x}}
\newcommand{\bz}{\mathbf{z}}
\newcommand{\bw}{\mathbf{w}}
\newcommand{\bX}{\mathbf{X}}
\newcommand{\bQ}{\mathbf{Q}}
\newcommand{\bV}{\mathbf{V}}

\newcommand{\bI}{\mathbf{I}}

\newcommand{\bbR}{\mathbb{R}}
\newcommand{\bbP}{\mathbb{P}}

\newcommand{\bSigma}{\boldsymbol{\Sigma}}
\newtheorem{remark}{Remark}
\acmJournal{POMACS}
\acmVolume{37}
\acmNumber{4}
\acmArticle{111}
\acmMonth{8}

\usepackage{algorithm}
\usepackage{algorithmic}

\begin{document}

\title{CICADA: Cross-Domain Interpretable Coding for Anomaly Detection and Adaptation in Multivariate Time Series}

\author{Tian Lan}
\email{lant23@mails.tsinghua.edu.cn}
\affiliation{%
	\institution{Tsinghua University}
	\city{BeiJing}
	\country{China}
}

\author{Yifei Gao}
\email{gao-yf@mail.tsinghua.edu.cn}
\affiliation{%
	\institution{Tsinghua University}
	\city{BeiJing}
	\country{China}
}

\author{Yimeng Lu}
\email{lu-ym21@mails.tsinghua.edu.cn}
\affiliation{%
	\institution{Tsinghua University}
	\city{BeiJing}
	\country{China}
}

\author{Chen Zhang}
\authornote{Corresponding author}
\email{zhangchen01@tsinghua.edu.cn}
\affiliation{%
	\institution{Tsinghua University}
	\city{BeiJing}
	\country{China}
}

\renewcommand{\shortauthors}{Trovato et al.}

\begin{abstract}
Unsupervised Time series anomaly detection plays a crucial role in applications across industries. However, existing methods face significant challenges due to data distributional shifts across different domains, which are exacerbated by the non-stationarity of time series over time. Existing models fail to generalize under multiple heterogeneous source domains and emerging unseen new target domains. To fill the research gap, we introduce CICADA (Cross-domain Interpretable Coding for Anomaly Detection and Adaptation), with four key innovations: (1) a mixture of experts (MOE) framework that captures domain-agnostic anomaly features with high flexibility and interpretability;  (2) a novel selective meta-learning mechanism to prevent negative transfer between dissimilar domains, (3) an adaptive expansion algorithm for emerging heterogeneous domain expansion, and (4) a hierarchical attention structure that quantifies expert contributions during fusion to enhance interpretability further. 
Extensive experiments on synthetic and real-world industrial datasets demonstrate that CICADA outperforms state-of-the-art methods in both cross-domain detection performance and interpretability.
\end{abstract}



\newcommand{\chen}[1]{{\color{red}{\bf{Chen says:}} \emph{#1}}}

\maketitle
\section{INTRODUCTION}
Time series anomaly detection aims to monitor sequential observations and identify those that deviate from the concept of normality. It plays a vital role in mission-critical applications across various domains, such as manufacturing \cite{li2021multivariate}, healthcare~\cite{pereira2019learning}, transportation \cite{van2019real}, cybersecurity~\cite{colo2019anomaly}, etc. In practice, the systems being monitored usually consist of multiple sensors and generate large volumes of multivariate time series, exhibiting complex spatiotemporal correlations that reflect the intricate dynamics of the system. We aim to detect any system-level anomaly, i.e., irregularities or unusual patterns of the multivariate time series. However, obtaining labels for system-level anomalies in multivariate time series is often challenging. It may not be easy to manually label each training data, especially the anomalies, and in many cases, it can be prohibitively difficult to even identify all potential anomaly types. Furthermore, the data is highly imbalanced, with anomalies being rare compared to the normal observations. 
As a result, there is an increasing trend toward leveraging unsupervised learning approaches for anomaly detection in multivariate time series, as they offer a more principled solution to handling the above challenges ~\cite{zamanzadeh2024deep}.

However, due to the natural evolution and nonstationarity of time series, both the data distribution and the underlying system dynamics often exhibit domain shifts. This means that the concepts of normality and anomalies can vary across different domains. For example, a manufacturing system can operate under different conditions, under each of which the distributions of normal data and abnormal data are different. As such, the anomaly detection model should be adaptable to different domains, This involves the following challenges. 

\begin{figure}[t]
\centering
\includegraphics[width=\linewidth]{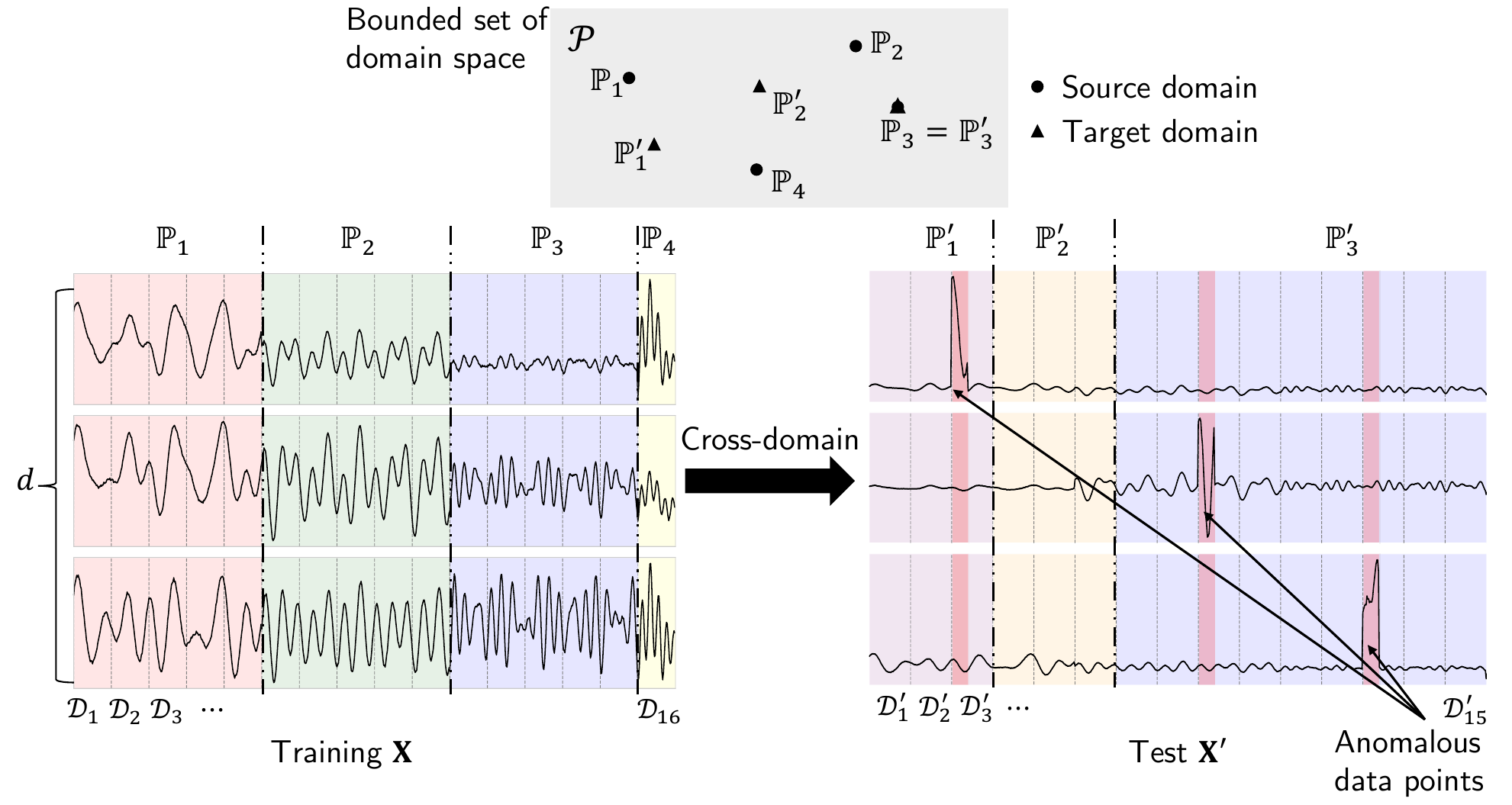}
\vspace{-25pt}
\caption{Cross-domain Anomaly Detection Problem}
\label{fig:problem}
\vspace{-20pt}
\end{figure}

First, both the training data and testing data may come from multiple source and target domains, as shown in Figure \ref{fig:problem}. While some studies have explored time series domain adaptation or generalization methods~\cite{ott2022domain, cai2021time, jin2022domain}, most of them typically assume a single source and a single target domain. 
Few studies have addressed multi-source domain adaptation \cite{wilson2023calda} or generalization \cite{deng2024domain}, which aim to leverage data from all the source domains to extract invariant features applicable to the target domain. However, this approach does not account for the potential heterogeneity between domains. Constructing a common feature space for heterogeneous domains may degrade the distinctive characteristics of dissimilar domains and result in \emph{negative transfer}. If different domains have heterogeneous or incompatible data features, say domain $\mathbb{P}_1$ and $\mathbb{P}_3$ in Figure \ref{fig:problem}, naively aggregating the features from all source domains may harm the anomaly detection performance~\cite{standley2020tasks,li2023few}. 

Second, domain shifts can be inherent in the nonstationary evolution of time series data and may be latent, making it difficult to precisely define the boundaries of each source or target domain. For example, in the manufacturing case, the system’s operating condition depends on factors such as the machine's health state and environmental settings, which can shift across different regimes. This inherent temporal nonstationarity leads to dynamic, unknown distribution shifts. We do not know which domain each time belongs to. So we call it as \textit{latent domain}. This necessitates automatic detection and continuous domain expansion for newly emerging latent source or target domains over time. This can be illustrated as adding points (each point represents a domain) in a bounded set of domain space we are interested in, denoted as $\calP$ ($\calP$ includes infinite points) in Figure \ref{fig:problem}. In the meanwhile, this adaptation must occur without catastrophic forgetting of previously learned domains. However, to date, no work has specifically addressed domain adaptation for time-varying and latent domains. In addition, existing domain adaptation methods in the literature typically assume only a single target domain. 

Third, anomaly features can be diverse, unpredictable, and domain-specific, making them difficult to capture consistently across different domains. It is therefore essential to make the model have strong \emph{flexibility and generalizability}, enabling it to effectively handle large-scale datasets and capture agnostic anomaly features as many as possible. In addition, the model must exhibit high interpretability, which is particularly critical in time series anomaly detection. Interpretability aids in system diagnostics and helps identify the root causes of anomalies, which is invaluable for downstream tasks.
 
To address the aforementioned challenges, we introduce \emph{CICADA} (Cross-domain Interpretable Coding for Anomaly Detection and Adaptation), a novel framework for unsupervised anomaly detection in multivariate time series that accounts for latent multi-source-and-target domain shifts. 
CICADA is built upon the mixture-of-experts (MOE) framework, and novelly proposes a selective meta-learning mechanism and integrates it with adaptive architecture expansion to enable robust cross-domain anomaly detection. The key components of CICADA include: 1) an attention-based multi-expert fusion mechanism that extracts the heterogeneous spatiotemporal features and agnostic anomaly patterns of multivariate time series, while also providing quantitative attribution for each expert to enhance interpretability; 2) 
a novel selective meta-learning mechanism that adapts each expert to any latent source or target domains, and avoids negative transfer from heterogeneous domains; 3) an adaptive expansion algorithm that dynamically detects emerging heterogeneous domains and expands each expert's capacity accordingly. 
Our principal contributions are as follows: 
\begin{itemize}[left = 0.1cm]
\item \textbf{Anomaly-agnostic detection:} 
CICADA leverages a MOE framework, where each expert represents certain interpretable data features that reflect certain anomaly patterns. These features can be derived from either machine learning-based or deep neural network-based models.

\item \textbf{Selective meta-learning}: To ensure adaptability across heterogenous domains, CICADA introduces a so-called selective meta-learning mechanism, that selects only similar (compatible) domains for adaptation, thereby preventing negative transfer from dissimilar (incompatible) domains.

\item \textbf{Adaptive expansion algorithm}: CICADA incorporates an automatic domain expansion algorithm that detects and integrates emerging new incompatible domains, further improving model generalization.

\item \textbf{Interpretable multi-expert Fusion}:  CICADA employs a hierarchical attention structure to dynamically weigh the experts and quantify their contributions, thus providing interpretable anomaly detection.

\item \textbf{Comprehensive validation}: Extensive experiments on synthetic and real-world datasets demonstrate CICADA's superior cross-domain anomaly detection performance compared to state-of-the-art baselines.
\end{itemize}

The remainder of this paper is organized as follows: Section \ref{sec:related-work} reviews the related work. Section \ref{sec:methodology} provides a detailed description of our CICADA architecture. Section \ref{sec:numerical} and Section \ref{sec:case} present experimental results on both synthetic and real-world datasets. Finally, Section \ref{sec:conclusion} offers conclusions and future research directions.

\section{RELATED WORK}
\label{sec:related-work}

\textbf{Unsupervised time-series anomaly detection:} The methods can be typically classified into four primary categories: prediction-based, reconstruction-based, clustering-based, and composite methods. Prediction-based methods construct predictive models and use prediction errors to detect anomalies over time~\cite{wu2022timesnet, xu2021anomaly}.
Reconstruction-based methods train models to learn and reconstruct normal patterns in the data, and identify anomalies as deviations from these reconstructed patterns \cite{malhotra2016lstm, audibert2020usad, su2019robust, kim2024model}. Clustering-based methods formulate anomaly detection as a one-class classification problem, learning latent representations that effectively distinguish normal from abnormal data \cite{darban2025carla, shen2020timeseries, xu2024calibrated, wang2023deep}. All the three categories above rely on constructing a single model, making them sensitive to domain shifts. In contrast, composite methods utilize a mixture of models, such as mixture encoders \cite{zhang2021unsupervised, han2022learning, zhao2021automatic} or prototype-oriented approaches \cite{li2023prototype} for prediction or reconstruction. Hence composite methods are more robust to domain shifts due to their ability to combine multiple models for more flexible anomaly detection.

\textbf{Domain adaptation and generalization in time-series analysis:} Domain adaptation aims to transfer knowledge from source domains to target domains. In time-series analysis, most existing domain adaptation methods focus on prediction or classification tasks \cite{ott2022domain, cai2021time, jin2022domain, he2023domain, wilson2023calda}. For anomaly detection, supervised approaches typically operate in a few-shot learning setting \cite{li2023few, lai2023context}, while unsupervised methods focus on feature alignment between source and target domains \cite{darban2024dacad, he2024variate,kim2024model}. However, these methods generally assume a single source and target domain, making adaptation more challenging when multiple source domains are involved.
Domain generalization, on the other hand, aims to train models that perform well on unseen domains without direct access to their data. In time-series analysis, existing domain generalization methods learn invariant features across multiple known domains and have been widely applied in forecasting \cite{deng2024domain} and classification \cite{wilson2023calda, zhang2022domain, lu2022out}. However, all the current studies do not address the issue of either negative transfer from incompatible domains or emerging incompatible domains, limiting their applicability in real-world scenarios. In addition, they assume which domain each time point belongs, i.e., the domain label, to be known in advance.

\textbf{Meta-learning for Domain Adaptation and Generalization:} Meta-learning is a crucial strategy for domain adaptation and generalization, as it focuses on acquiring meta-knowledge that can be transferred across different tasks (domains). In domain adaptation, meta-learning methods aim to adapt to new tasks with unlabeled or insufficient data by transferring meta-knowledge \cite{chen2019blending, feng2021similarity}. In domain generalization, meta-learning learns an initial set of parameters that can be fine-tuned through multiple gradient steps to adapt to new tasks \cite{finn2017model}. 
In particular, Model-Agnostic Meta-Learning (MAML) is one of the most widely used approaches due to its bi-level optimization framework \cite{li2019feature, li2018learning, shu2021open}. However, very few studies address challenges related to domain heterogeneity and negative transfer \cite{zhao2024more}. Moreover, there is currently no research that leverages meta-learning for domain adaptation or generalization specifically in the context of time-series anomaly detection. Last, existing methods also assume the domain label of each observation to be known in advance.

\textbf{Task clustering for negative transfer:} 
To address task heterogeneity, hierarchically structured meta-learning has been proposed to reduce negative transfer and improve performance across various tasks \cite{yao2019hierarchically}. This approach utilizes a hierarchical task clustering structure to group tasks and tailor transferable knowledge to different clusters. Task clustering can be achieved either based on task similarities \cite{yao2019hierarchically, liu2023hierarchical} or by minimizing validation loss \cite{standley2020tasks, fifty2021efficiently, song2022efficient}. These methods are primarily designed for classification tasks with known task labels. In our context, since the number of domains and the domain labels are unknown, clustering tasks (domains) may be inapplicable.

\begin{figure*}[t]
    \centering
    \includegraphics[width=1\textwidth]{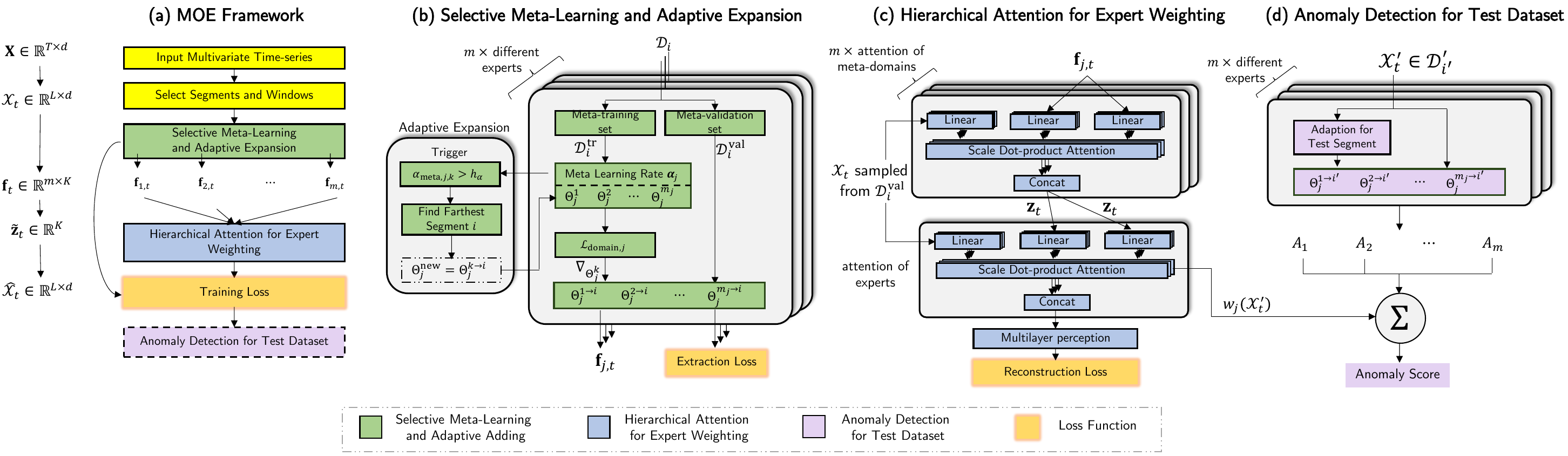}
    \vspace{-20pt}
    \caption{The framework of CICADA}
    \vspace{-13pt}
    \label{fig:framework}
\end{figure*}

\section{METHODOLOGY}
\label{sec:methodology}
\textbf{Problem formulation:} Consider a multivariate time series training dataset \(\bX = (\bx_1, \bx_2, \ldots, \bx_T) \in \mathbb{R}^{T \times d}\), where \(T\) denotes the length of the time series, \(d\) denotes the number of variables, and \(\bx_t\) is the observation at time step \(t\). We assume the $\bX$ belong to multiple source domains $\bbP_1,\bbP_2,\ldots$, which are in a bounded set of domain space, denoted as $\calP$. Given a new time series testing dataset \(\bX'=(\bx_1',\bx_2',\ldots, \bx_{T'}') \in \mathbb{R}^{T' \times d}\) that belongs to multiple target domains $\bbP_1',\bbP_2',\ldots$, also in $\calP$, our goal is to detect anomalous observations in $\bX'$, and output a label vector \(\by' = (y_1', y_2', \ldots, y_{T'}')\), where \(y_t' \in \{0, 1\}\) indicates the presence (1) or absence (0) of an anomaly at \(t\). 

To account for temporal correlations of $\bx_t$, we select a window size \( L \), and use \( \mathcal{X}_t = (\mathbf{x}_{t-L}, \mathbf{x}_{t-L+1}, \ldots, \mathbf{x}_{t})^\top \in \mathbb{R}^{L \times d} \) in the window for modeling and anomaly detection analysis at each $t$. The training data and testing data can then be reformulated as \( \mathcal{X} \in \mathbb{R}^{T \times L \times d} \) and \( \mathcal{X'} \in \mathbb{R}^{T' \times L \times d} \). Since the data domain are expected to shift over time, we partition $\mathcal{X}$ into different time segments \( \mathcal{D}_1, \mathcal{D}_2, \ldots, \mathcal{D}_N \), where each segment corresponds to a time interval $\calD_i=\{\calX_{t}|t_i\leq t<t_{i+1}\}$. We assume that each $\calD_i$ belongs to a latent source domain $\mathbb{P}_{d_i} \in \calP$, and different $\mathbb{P}_{d_i}$ may be same, similar or distinct from one another. For the testing data, we apply the same approach to partition $\mathcal{X}'$ into distinct segments \( \mathcal{D}_1^{\prime}, \mathcal{D}_2^{\prime}, \ldots, \mathcal{D}_{N'}^{\prime}\) where $\calD_{i^{\prime}}^{\prime}=\{\calX_{t}^{\prime}|t'_i\leq t<t'_{i+1}\}$. We assume $\mathcal{D}_{i^{\prime}}^{\prime}$ also belongs to unknown latent target domain $\bbP_{d_i^{\prime}}^{\prime} \in \calP$.

\subsection{Mixture-of-Experts (MOE) Framework}

We assume CICADA consists of \( m \) experts, with each expert corresponding to a specific interpretable feature extraction method \( g_j \) and its parameters
denoted as \( \Theta_j \). To capture the differences and similarities of \( \Theta_j \) across different domains, for expert $j$, we introduce $m_j$ meta-domains $\calP_j^1,\calP_j^2,\ldots,\calP_j^{m_j}$ as a division for $\calP$, and we suppose these meta-domains are distinct, i.e., $\calP_j^k \cap \calP_j^{k'}=\emptyset,\forall k\ne k^{\prime}$, which can be shown in Figure \ref{fig:meta-learning}(b). The meta-domains of different experts are independently defined, and hence their divisions of $\calP$ are different, but all satisfy $\cup_{k=1}^{m_j}\calP_j^k=\calP, \forall j = 1,\ldots, m$. For each expert $j$, each meta-domain $\calP_{j}^{k}$ has an initial model parameter $\Theta_j^{k}$, and $\Theta_j^{k}$ for $ k=1,\ldots, m_j$ are far apart from each other. 

For each $\calD_{i}$(or $\calD_{i^{\prime}}^{\prime}$), its associated domain $\bbP_{d_i}$(or $\bbP'_{d_i^{\prime}}$) can be selectively adapted from one of the $m_j$ basic meta-domains. Specifically, we define an indicator $\delta_{j}^{k \to i} (\text{or } \delta^{k\rightarrow i^{\prime}}_{j}) \in \{0,1\}$ satisfying $\sum_k\delta_j^{k\rightarrow i}=1,\forall j=1,2,\ldots,m$. If $\delta_{j}^{k \to i} (\text{or } \delta^{k\rightarrow i^{\prime}}_{j})=1$, it indicates that $\bbP_{d_i}$(or $\bbP'_{d_i^{\prime}}$)$\in\calP_j^k$, and its initial parameters $\Theta_j^{k}$ can be adapted to $\calD_{i}$(or $\calD_{i^{\prime}}^{\prime}$), resulting in an adapted model parameter as $\Theta_j^{k \rightarrow i}$(or $\Theta_j^{k \rightarrow i^{\prime}}$). Otherwise, $\delta_{j}^{k \to i} (\text{or } \delta^{k\rightarrow i^{\prime}}_{j}) =0$. As such, if two $\mathcal{D}_{i}$(or $\mathcal{D}_{i^{\prime}}^{\prime}$) belong to the same meta-domain, they are considered similar. Conversely, if they belong to different meta-domains, they are deemed dissimilar. In this way, we can capture the heterogeneity of domains. 

It is important to note that since the meta-domains of different experts are independently defined, the meta-domain a $\mathcal{D}_{i}$ lies, i.e., $k_j = \arg\max_{k}\delta^{k\rightarrow i}_{j}$ may differ for different experts. This is reasonable, as the domain shifts can manifest differently for different experts. Hence for a domain $\bbP_{d_i}$, its meta-domain label can be given by $[k_1,k_2,\ldots,k_m]$.

To detect anomalous observations in $\calX^{\prime}$, for any $\calX_{t}^{\prime}\in \mathcal{D}_{i^{\prime}}^{\prime}$, we first get its adapted model parameters \( \Theta_j^{k \rightarrow i^{\prime}} \), and then extract the corresponding extracted feature as follows:
\[
\mathbf{f}_{j,t}^{k'} = g_j(\mathcal{X}_t^{\prime}; \Theta_j^{k \rightarrow i^{\prime}}), \forall \mathcal{X}_t^{\prime} \in \mathcal{D}_{i^{\prime}}^{\prime}.
\]
We also define another anomaly score function $A_{j}(\mathcal{X}_{t}^{\prime}; \Theta_{j}^{k \to i^{\prime}})$ and compute the final anomaly score as 
\begin{equation}
\label{equ:anosc}
\text{anosc}(\mathcal{X}'_t) = \sum_{j=1}^m \sum_{k=1}^{m_j} w_j(\mathcal{X}'_t) \delta_j^{k \rightarrow i^{\prime}} A_j(\mathcal{X}'_t; \Theta_j^{k \rightarrow i^{\prime}}), \forall \mathcal{X}'_t \in \calD_{i^{\prime}}^{\prime}.
\end{equation}
Here $w_j(\mathcal{X}'_t)$ represents the weight of expert $j$'s contribution. If the anomaly score $\text{anosc}(\mathcal{X}'_t)$ exceeds a predefined threshold, we classify $\mathcal{X}'_t$ as anomaly.

Section \ref{sec:selective meta-learning} introduces our selective meta-learning framework to learn $\Xi_j=\{\Theta_{j}^{1},\ldots,\Theta_{j}^{m_j}\}$ from $\calX$ given a fixed $m_j$. This allows the $m_j$ meta-domains have distinct parameters that can capture the features from heterogeneous domains. For each target $\calD_{i^{\prime}}^{\prime}$, only one meta-domain is selected to learn $\Theta_{j}^{k\to i^{\prime}}$, thereby avoiding negative transfer. Section \ref{sec:adaptive_add} describes how to increase the number of meta-domains, and selects the appropriate \( m_j \). Initially, we assume \( \Xi_j = \{\Theta_j^1\} \) with a single meta-domain, and progressively generate new meta-domains, stopping this expansion when we have sufficient meta-domains. Section \ref{sec:feature_fusion} discusses how to fuse the anomaly scores from different experts, i.e., how to estimate $w_j(\calX_{t}^{\prime})$. Section \ref{sec:train} summarizes the training process. 
Last, Section \ref{sec:test_adaptation} presents the adaptation process for the testing data $\calX^{\prime}$, and describe the overall anomaly detection scheme. The entire CICADA framework is summarized in Figure \ref{fig:framework}.

\subsection{Selective Meta-Learning}
\label{sec:selective meta-learning}
We begin by assuming that the number of meta-domains is known, and define the set of initial parameters of all the meta-domains as \( \Xi_j=\{\Theta_{j}^{1},\ldots, \Theta_{j}^{m_j}\} \). We then propose a novel selective meta-learning method by improving the first-order MAML~\citep{finn2017model} to allow heterogeneous tasks, to learn $\Xi_j$ using the training data $\calX$. Specifically, for each $\calD_{i}$, we randomly split it into a meta-training set \( \mathcal{D}_i^{\rm tr} \) and a meta-validation set \( \mathcal{D}_i^{\rm val} \). 

\textbf{Selective domain-specific update:} 
During the selective domain-specific update, each expert $j$ attends to adapt the meta-domain with parameter $\Theta_j^k$ to $\calD_{i}$ according to selective domain-specific update as: 
\begin{equation}
\Theta_j^{k \rightarrow i} = \Theta_j^k - \alpha_{{\rm meta},j,k} \nabla_{\Theta_j^k} \mathcal{L}_{\text{domain},j}(\mathcal{D}^{\rm tr}_i; \Xi_j),
\end{equation}
where \( \alpha_{{\rm meta},j,k} \) is the meta-learning rate, a learnable parameter that enables adaptive expansion (further details on this process are discussed in Section \ref{sec:adaptive_add}). $\mathcal{L}_{\text{domain},j}(\mathcal{D}^{\rm tr}_i; \Xi_j)$ is defined as 
\begin{equation}
\label{eq: meta_train}
 \mathcal{L}_{\text{domain},j}(\mathcal{D}^{\rm tr}_i; \Xi_j)=\min_{k=1,\ldots, m_j} \mathcal{L}_j(\mathcal{D}^{\rm tr}_i; \Theta_j^{k}).
\end{equation}
Here \( \mathcal{L}_{j}(\mathcal{D}^{\rm tr}_i; \Theta_j^k) \) is the optimization objective for the feature extraction method of expert $j$, which could be either machine learning-based or deep learning-based. Examples of $\mathcal{L}_j(\mathcal{D}^{\rm tr}_i; \Theta_j^k)$ for some commonly used experts are provided in Appendix \ref{app:anomaly_score_function}.

(\ref{eq: meta_train}) implies that we select the meta-domain with the initial parameters $\Theta_{j}^{k}$ that minimize $\mathcal{L}_{j}$, i.e., $k^{*} = \arg\min_{k=1,\ldots, m_j} \mathcal{L}_{j}(\mathcal{D}^{\rm tr}_i; \Theta_j^k)$, and only conduct adaptation from the meta-domain $k^{*}$. While for the other meta-domains $k\neq k^{*}$, no adaptation occurs since $\nabla_{\Theta_j^k} \mathcal{L}_{\text{domain},j}(\mathcal{D}^{\rm tr}_i; \Xi_j)=\mathbf{0}$. This update provides two key advantages: specialization and selective adaptation. Specialization allows expert $j$ to tailor \( \Theta_j^{k \rightarrow i} \) to the specific meta-domain represented by \( \mathcal{D}_i^{\rm tr} \), while selective adaptation prevents unnecessary or unreasonable adaptations from other dissimilar meta-domains. 

\textbf{Selective meta-domain update:} After updating the model parameters for each domain, we evaluate the adapted model with parameters \( \Theta_j^{k \rightarrow i} \) on the validation set $\calD_{i}^{\text{val}}$, and update the initial parameters $\Xi_{j}$ for all the meta-domains to minimize the following loss:
\begin{equation}
\label{eq:meta_test}
\mathcal{L}_{\text{meta}, j}(\mathcal{D}^{\rm val}_i;\Xi_j) = \sum_{k=1}^{m_j} \delta^{k\rightarrow i}_{j} \cdot \mathcal{L}_j(\mathcal{D}^{\rm val}_i; \Theta_j^{k \rightarrow i}).
\end{equation}
Here $\delta^{k\rightarrow i}_{j} = \mathbb{I}(k=\underset{k=1,\ldots,m_j}{\arg \min}(\calL_j(\calD^{\rm tr}_i;\Theta_j^k)))$. 

To ensure that the initial parameters $\Theta_j^k$ of each meta-domain are closed to the corresponding adapted parameters of each domain, we add regularization terms to penalize the meta-learning rates $\boldsymbol{\alpha}_j=\{\alpha_{\text{meta},j,1},\ldots,\alpha_{\text{meta},j,m_j}\}$. The penalty term can be expressed as $\calL_{\text{pen},j}(\boldsymbol{\alpha}_j)= \sum_{k} \alpha_{\text{meta},j,k}$.
Finally, we define the extraction loss based on our selective meta-learning as:
\begin{equation}
\calL_{\text{extraction}} =\sum_{i=1}^N \sum_{j=1}^m\calL_{\text{meta},j}(\calD_i^{\rm val}; \Xi_j)+ \sum_{j=1}^m \lambda_{\text{pen}} \calL_{\text{pen},j}(\boldsymbol{\alpha}_j).
\end{equation}

\begin{figure}
    \centering
    \includegraphics[width=\linewidth]{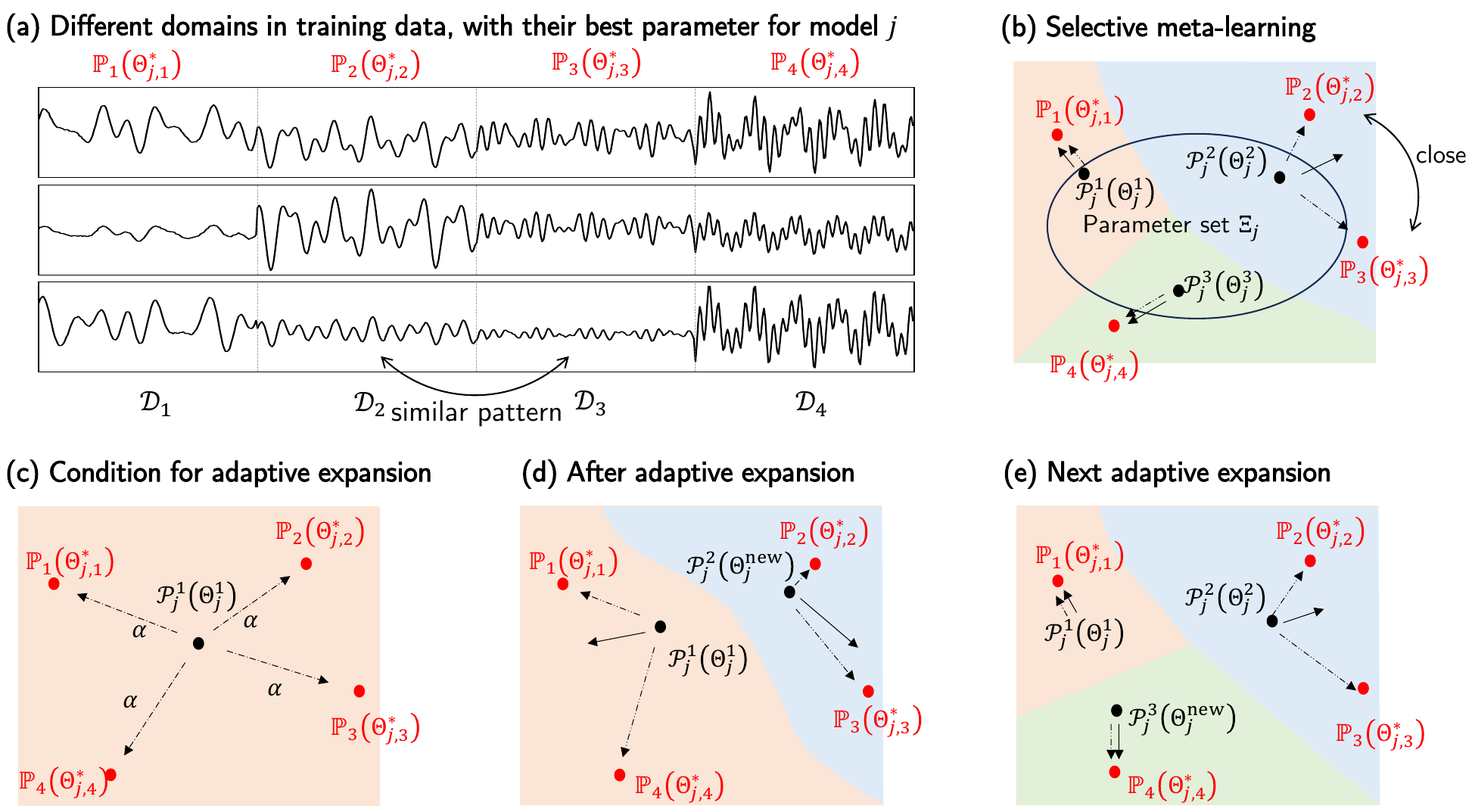}
    \vspace{-20pt}
    \caption{(a): Time series with multiple domains; (b): The idea of meta-domain and selective meta-learning; (c) to (e): The adaptive expansion process (dotted lines represent adaptation directions and solid lines represent movement directions of meta-domains' initial parameters).}
    \vspace{-16pt}
    \label{fig:meta-learning}
\end{figure}

\textbf{Illustration of preventing negative transfer:} 
To illustrate how the selective meta-learning works, consider the example shown in Figure \ref{fig:meta-learning}. Suppose we have a time-series training dataset divided into four segments $\calD_{1},\calD_{2}, \calD_{3}, \calD_{4}$, corresponding to domains \( \mathbb{P}_1, \mathbb{P}_2, \mathbb{P}_3, \mathbb{P}_4 \), where $d_i=i$ for each domain. Each domain \( \mathbb{P}_i \) has its optimal parameters for expert \( j \), denoted as:
\begin{equation}
    \Theta_{j,i}^* = \underset{\Theta_{j,i}}{\arg \min}\ \mathbb{E}_{\mathcal{X}_t \sim \mathbb{P}_i} \left[ \mathcal{L}_j(\mathcal{X}_t; \Theta_{j,i}) \right].
\end{equation}
Assume that we have three meta-domains $\calP_{j}^{1}, \calP_{j}^{2}, \calP_{j}^{3}$, and their current initial parameters are \( \Xi_j = \{\Theta_j^1, \Theta_j^2, \Theta_j^3\} \) respectively, as shown in Figure \ref{fig:meta-learning}(b).
In the selective domain-specific update, we compute the domain-specific adaptation indicators $\delta_{j}^{k\to i}$. We get $\delta_{j}^{1\to 1}=1, \delta_{j}^{2\to 2}=1, \delta_{j}^{2\to 3}=1, \delta_{j}^{3\to 4}=1$, indicating that $\mathbb{P}_1$ belongs to the meta-domain $\calP_{j}^{1}$ with $\Theta_{j}^{1}$; 
\( \mathbb{P}_2 \) and \( \mathbb{P}_3 \) belong to meta-domain $\calP_{j}^{2}$ with $\Theta_{j}^{2}$, and $\mathbb{P}_4$ belongs to the meta-domain $\calP_{j}^{3}$ with $\Theta_{j}^{3}$ respectively. 

Consequently, when adapting to $\mathbb{P}_1$, we select the meta-domain $\calP_{j}^{1}(\Theta_j^1)$ and avoid the negative transfer from the other two meta-domains. Similarly, when adapting to $\mathbb{P}_i, i = 2,3,4$, we avoid selecting the meta-domain $\calP_{j}^{1}(\Theta_j^1)$, thereby preventing negative transfer. As a result, for the selective meta-domain update, the update of \( \Theta_j^1 \) is driven solely by the data from $\mathbb{P}_1$, moving it toward the optimal parameters $\Theta_{j,1}^{*}$. 

For domains $\mathbb{P}_2$ or $\mathbb{P}_3$, we select the meta-domain $\calP_{j}^{2}(\Theta_j^2)$ and prevent negative transfer from the other two meta-domains. Consequently, during the selective meta-domain update, the update of \( \Theta_j^2 \) are influenced by the data from both $\mathbb{P}_2$ and $\mathbb{P}_3$, making it move closer to $\Theta_{j,2}^{*}$ and $\Theta_{j,3}^{*}$. 

The same process applies to $\mathbb{P}_4$ and $\calP_{j}^{3}(\Theta_j^3)$. In summary, the expert effectively learns domain-specific parameters and meta-domain initial parameters without negative transfer from dissimilar domains, respectively. 

\subsection{Adaptive Expansion}
\label{sec:adaptive_add}
Next, we discuss how to determine the number of meta-domains for each expert, i.e., \( |\Xi_j| \) and how to adaptively add a new meta-domain with initial parameters \( \Theta_{j}^{\text{new}} \) to \( \Xi_j \) during the training process. 

\textbf{Cases requiring meta-domain expansion:} 
We begin from the general case with \( \Xi_j = \{\Theta_j^1,\Theta_j^2,\ldots,\Theta_j^{m_j}\} \). After completing selective meta-learning with the current $m_j$ meta-domains, we assume theoretically we have reached a local optimum for $\Theta_{j}^{k}$, i.e., 
\begin{equation}
\label{equ:local-optim}
\nabla_{\Theta_j^k} \sum_{\mathbb{P}_i \sim p(\mathbb{P})} \mathbb{E}_{\mathcal{X}_t \sim \mathbb{P}_i} \left[ \delta_j^{k \rightarrow i} \cdot \mathcal{L}_{j}(\mathcal{X}_t; \Theta_{j}^{k \rightarrow i}) \right] = \mathbf{0}.
\end{equation}

In an ideal scenario, if $m_j$ meta-domains are sufficient,  \( \Theta_{j}^{k}\) is expected to be close to \( \Theta_{j,i}^* \) for all $i$ where $\delta_j^{k \rightarrow i}=1$. In this case, the adaptation meta-learning rate $\alpha_{\text{meta}, j, k}$ used to update \( \Theta_{j}^{k\to i}\) would be small. However, if the $m_j$ meta-domains are insufficient, and the optimal parameters \( \Theta_{j,i}^* \) for all $i$ satisfying $\delta_j^{k \rightarrow i}=1$ are significantly dispersed, achieving good adaptation would require a larger optimal meta-learning rate \( \alpha_{\text{meta},j,k} \).

In this case, large differences between domains within the same meta-domain lead to an increased optimal \( \alpha_{\text{meta},j,k} \). This indicates unreasonable adaptations from the meta-domains. In other words, the current set of meta-domains is insufficient to adapt the parameters effectively to all the domains, signaling the need for an expansion of the meta-domains.

\textbf{Adaptive expansion algorithm:} We treat \( \alpha_{\text{meta},j,k} \) as a learnable parameter and define a threshold $h_{\alpha}$ to decide whether a new meta-domain should be added. Specifically, if the condition
\begin{equation}
\label{equ:adaptive-add}
\alpha_{\text{meta},j,k} > h_{\alpha}
\end{equation}
is met, we update the set of parameters \( \Xi_j \) by introducing a new meta-domain $\calP_{j}^{\text{new}}$ with initial parameters \( \Theta_j^{\text{new}} \) defined as:
\[
    \Theta_j^{\text{new}} = \Theta_j^{k \rightarrow i_{\text{farthest}}}  \text{ where } i_{\text{farthest}} = \underset{\{i=1,\ldots, N|\delta_{j}^{k\to i}=1\}}{\arg \max} \ \| \Theta_j^{k \rightarrow i} - \Theta_j^k \|_F^2.
\]
In the case that multiple $k$ satisfy the condition in \eqref{equ:adaptive-add}, we can select the $k$ with maximum meta-learning rate to perform the adaptive expansion.

\begin{remark}
After introducing $\Theta_j^{\rm new}$, it inhibits the previous meta-domain  \(\calP_{j}^{k}(\Theta_j^{k})\) to adapt to domain $i_{\text{farthest}}$, causing $\delta_{j}^{k \to i_{\text{farthest}}}$ to become 0. This results in a disruption of the balance in \eqref{equ:local-optim}, i.e., 
\begin{equation}
\nabla_{\Theta_j^k} \sum_{\mathbb{P}_i \sim p(\mathbb{P})} \mathbb{E}_{\mathcal{X}_t \sim \mathbb{P}_i} \left[ \delta_{j}^{k \rightarrow i} \cdot \mathcal{L}_{j}(\mathcal{X}_t; \Theta_{j}^{k \rightarrow i}) \right]\ne \mathbf{0}.
\end{equation}
Consequently, \( \Theta_j^k \) escapes the local optimum, as shown in ~\eqref{equ:local-optim} and continues updating. Since it requires several epochs for the parameters to converge to the new optimal solution after the introduction of $\Theta_j^{\rm new}$, (\ref{equ:adaptive-add}) is evaluated only at fixed epochs to prevent overfitting or premature updates. 
\end{remark}


\textbf{Illustration of adaptive expansion:} We utilize the example shown in Figures \ref{fig:meta-learning}(c) to (e) to illustrate the operation of adaptive expansion. Initially, there exists only one meta-domain with $\Xi=\{\Theta_j^1\}$, which will lead to significantly large adaptive meta-learning rate $\alpha_{{\rm meta},j,1}$. After the first adaptive expansion, a new meta-domain is introduced, encompassing domains $\bbP_2$ and $\bbP_3$, with initial parameters $\Theta_j^2=\Theta_j^{\rm new}$. As a result, $\Theta_j^1$ converges closer to the optimal parameters $\Theta_{j,1}^*$ and $\Theta_{j,4}^*$, leading to a reduction in its meta-learning rate. However, despite achieving local optimality, the meta-learning rate remains above the threshold. Subsequently, upon the next adaptive expansion, a new meta-domain is further added with parameter $\Theta_j^3=\Theta_j^{\text{new}}$. As a result, we obtain three meta-domains with initial parameters $\Xi_j=\{\Theta_j^1,\Theta_j^2,\Theta_j^3\}$.

\textbf{Evolutionary insight for selective meta-learning and adaptive expansion:} We provide an interesting evolutionary insight for our selective meta-learning and adaptive expansion mechanisms by cicadas (a type of insect) in Appendix \ref{app:evolutionary-insight}, demonstrating the essential rationality of these two mechanisms.

\subsection{Hierarchical Attention for Expert Weighting}
\label{sec:feature_fusion}
Based on the meta-domains learned with $\Xi_{j}$, in this section, we discuss how to assign weights $w_{j}(\cdot)$ in \eqref{equ:anosc} to each expert $j=1,\ldots, m$. In our unsupervised setting, experts are weighted according to their ability to reconstruct the data. The key idea is that the more an expert's features contribute to data reconstruction, the more weight it should receive for anomaly detection. This can be achieved via the following hierarchical attention structure. 

\textbf{Attention of meta-domains:} For each expert, we first fuse the features from its different meta-domains. Specifically, for each $\calX_{t}\in \calD_{i}^{\text{val}}$, the feature extracted from the meta-domain $\calP_{j}^{k} (\Theta_{j}^{k})$ is given by $\mathbf{f}_{j,t}^k = g_j(\calX_t; \Theta_j^{k \rightarrow i}) \in \mathbb{R}^{K}$. All the features are then concatenated as $\mathbf{f}_{j,t} = (\mathbf{f}_{j,t}^1, \mathbf{f}_{j,t}^2, \ldots, \mathbf{f}_{j,t}^{m_j}) \in \mathbb{R}^{K\times m_j}$. 

We employ multi-head attention to fuse the features from different meta-domains. For each attention head \( l \in \{1, 2, \ldots, h\} \), we have 
\begin{align*}
\bQ_{l,j} = \bW^Q_{l,j} \, \text{vec}(\mathcal{X}_t), \quad \bK_{l,j} = \bW^K_{l,j} \, \mathbf{f}_{j,t}, \quad \bV_{l,j} = \bW^V_{l,j} \, \mathbf{f}_{j,t}, \\ 
\bH_{l,j}=\text{Attention}(\bQ_{l,j}, \bK_{l,j}, \bV_{l,j}) = \text{softmax}\left( \frac{\bQ_{l,j}^\top\bK_{l,j}}{\sqrt{K/h}} \right) \bV_i^\top,
\end{align*}
where $\text{vec}(\mathcal{X}_t)\in \mathbb{R}^{Ld}$ is the vectorized form of $\calX_{t}$, and the weight matrics \( \bW^Q_{l,j} \in \mathbb{R}^{(K/h) \times Ld} \), \( \bW^K_{l,j} \in \mathbb{R}^{(K/h) \times K} \), and \( \bW^V_{l,j} \in \mathbb{R}^{(K/h) \times K} \) are learnable parameters. The softmax function is applied across the \( m_j \) meta-domains to generate attention weights that sum to one, producing $\bH_{l,j}\in \mathbb{R}^{K/h}$. 

We concatenate the attention outputs from all $h$ heads, and apply a linear transform $\bW_{j}^O$:
\begin{equation*}
\bz_{j,t} = \bW^O_j\text{concat}(\bH_{1,j}, \ldots,\bH_{h,j}),
\end{equation*}
where \( \bW^O_j \in \mathbb{R}^{K \times K} \) is a learnable parameter, and $\bz_{j,t} \in \mathbb{R}^{K}$ is the fused feature of expert $j$. All the fused features are concatenated as $\bz_{t}=(\bz_{1,t},\ldots,\bz_{m,t})\in \mathbb{R}^{K\times m}$. 

\textbf{Attention of experts:} Next, we fuse $\bz_t$ to obtain the final fused feature $\tilde{\bz}_t \in \bbR^K$ through expert-level multi-head attention, following a similar approach as the attention of meta-domains. Specifically, for each head $l = 1,\ldots, \tilde{h}$,
\begin{equation*}
\begin{split}
\tilde{\bQ}_l &= \tilde{\bW}^Q_l {\rm vec}(\calX_t), \tilde{\bK}_{l} = \tilde{\bW}^K_{l} \, \mathbf{z}_t, \bV_{l} = \tilde{\bW}^V_l \mathbf{z}_{t}, \\
\tilde{\bH}_l&={\rm Attention}(\tilde{\bQ}_l,\tilde{\bK}_l,\tilde{\bV}_l) = \text{softmax}\left( \frac{\tilde\bQ_{l}^\top\tilde\bK_{l}}{\sqrt{K/h}} \right) \tilde\bV_l^\top, 
\end{split}
\end{equation*}
 where $\tilde{\bW}^Q_l \in \mathbb{R}^{(K/h) \times Ld}$, $\tilde{\bW}^K_{l} \in \mathbb{R}^{(K/h) \times K}$ and $\tilde{\bW}^V_l \in \mathbb{R}^{(K/h) \times K}$. We can assign a weight to each expert $\bw(\calX_t)=(w_1(\calX_t), \ldots, w_m(\calX_t))$ in (\ref{equ:anosc}) by the attention score: 
\begin{equation}
   \bw(\calX_t)=\frac{1}{\tilde{h}}\sum_{l=1}^{\tilde{h}}\left(\text{softmax}\left( \frac{\tilde\bQ_l \tilde\bK_l}{\sqrt{K/h}} \right)\right).
\end{equation}
In the meanwhile, we concatenate the attentions of all the $\tilde{h}$ heads and apply a linear transform $\tilde{\bW}^{O}$ to get the final fused feature representation: 
\begin{equation*}
\tilde{\bz}_t=\tilde{\bW}^O {\rm concat}(\tilde{\bH}_1,\tilde{\bH}_2,\ldots,\tilde{\bH}_h).
\end{equation*}
Finally, we pass $\tilde{\bz}$ through a multi-layer perception to reconstruct $\calX_{t}$. The reconstruction loss is then computed as:  
\begin{equation}
    \calL_{\text{reconstruction}} = \sum_{i=1}^{N}\sum_{\calX_t\in \calD_{i}^{\rm val}}\| \calX_t - \text{MLP}(\tilde{\bz}_t; \Theta_{\text{MLP}})\|_F^2.
\end{equation}
\subsection{Training Process}
\label{sec:train}
We train the parameters $\Xi_{j}$, $\boldsymbol{\alpha}_j$, $\bW^Q_{l,j}$, $\bW^K_{l,j}$, $\bW^V_{l,j}$ for $j=1,\ldots, m, l = 1,\ldots, h$, $\tilde{\bW}^Q_l$, $\tilde{\bW}^K_{l}$ and $\tilde{\bW}^V_l$ for $l = 1,\ldots, \tilde{h}$, as well as $\Theta_{\text{MLP}}$ based on the training data $\calX$ using the following overall loss function: 
\begin{equation}
\calL=\calL_{\text{reconstruction}} + \lambda_1\calL_{\text{extraction}},
\end{equation}
where $\lambda_1$ is the hyperparameter balancing the two parts. When $\lambda_1$ is large, CICADA prioritizes the feature extraction $\mathbf{f}_{j,t}$ based on the original loss function for each expert. When $\lambda_1$ is small, experts are more heavily influenced by the reconstruction loss. The detailed training algorithm for CICADA is provided in Algorithm \ref{alg:CICADA} in Appendix \ref{app:algorithms}.
\subsection{Testing Process}
\label{sec:test_adaptation}
For a new time segment $\calD_{i}^{\prime}\in \calX^{\prime}$, each expert \(j\) first adapts the initial parameters of each meta-domain $\Theta_{j}^{k}$ to the a new data segment as follows:
\[
\Theta_j^{k \rightarrow i^{\prime}} = \Theta_j^k - \alpha_1 \nabla_{\Theta_j^k} \mathcal{L}_{\text{domain},j}(\calD'_{i^{\prime}}; \Xi_j),
\]
where \(\alpha_1\) is the fixed meta-learning rate for the testing dataset, treated as a hyperparameter. 
Next we compute its anomaly score for $\calX_{t}^{\prime}\in \calD_{i^{\prime}}^{\prime}$ based on \eqref{equ:anosc}. 
   \begin{figure}[t]
    \centering
    \includegraphics[width=\linewidth]{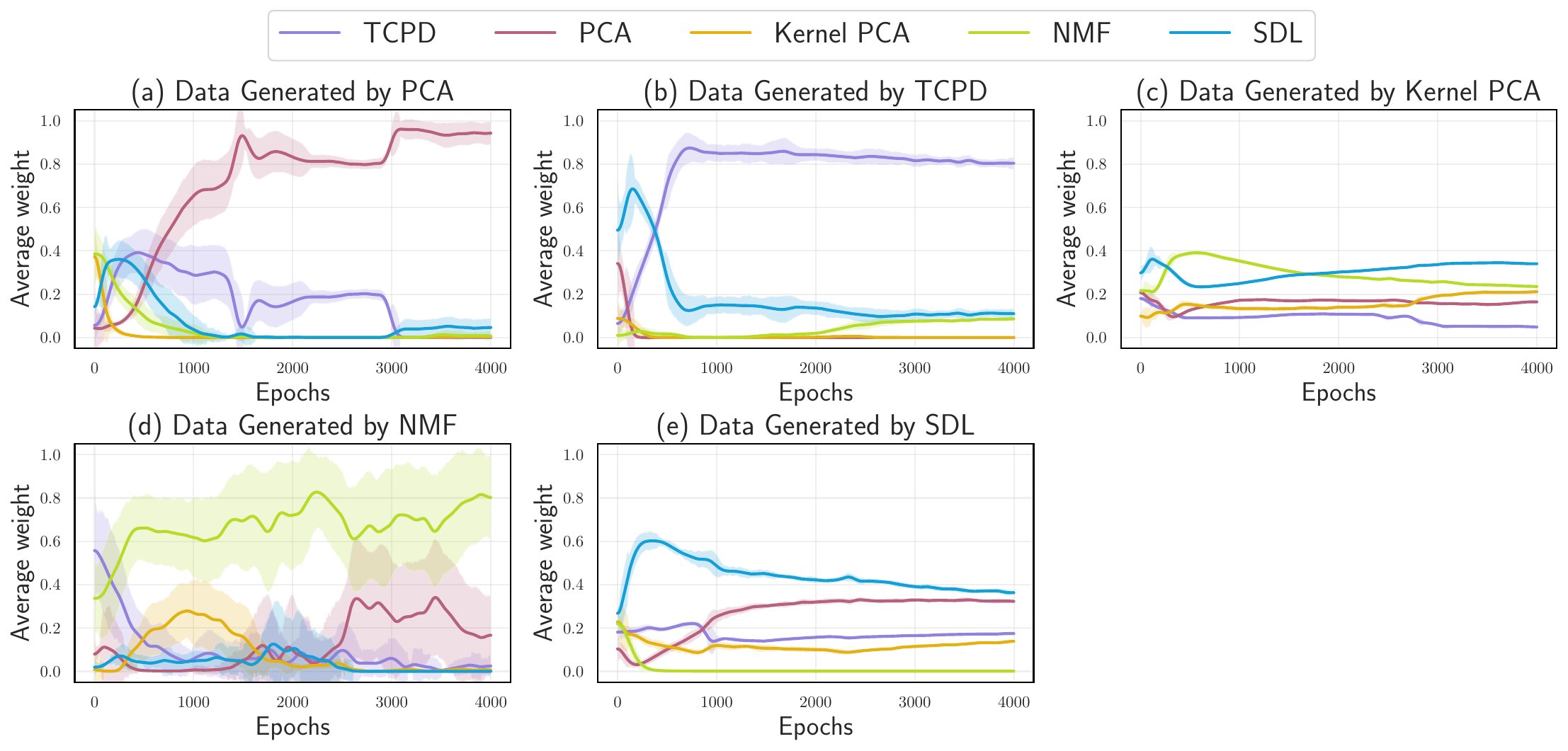}
    \vspace{-23pt}
    \caption{Average expert weight in the CICADA training process.}
    \vspace{-15pt}
    \label{fig:interpretable weight}
\end{figure}
Appendix \ref{app:anomaly_score_function} provides commonly used anomaly scores for different experts employed in this paper. 
Finally, by setting a threshold $C$ on the anomaly score, we predict the anomaly label \(y_t'\) as follows:
\[
y_t'=
\begin{cases}
0 & \text{if } \text{anosc}(\mathcal{X}_t') < C, \\
1 & \text{if } \text{anosc}(\mathcal{X}_t') > C.
\end{cases}
\]
\section{NUMERICAL STUDIES}
\label{sec:numerical}
We design two synthetic experiments to evaluate CICADA's ability to produce interpretable expert weights and to test its capability to adaptively add different meta-domains.
\subsection{Experiments for Interpretable Expert Weight}
\label{sec:numerical_1}
The objective of this experiment is to demonstrate the efficiency of our MOE framework and its interpretability that can identify the most suitable expert for a given dataset. In particular, we configure CICADA with five machine learning experts, i.e., $m=5$, with \( g_1 \) to \( g_5 \) representing Principal Component Analysis (PCA), Kernel PCA, Nonnegative Matrix Factorization (NMF), Tensor CP Decomposition (TCPD), and Sparse Dictionary Learning (SDL). For each expert, we generate synthetic time-series data \( \bX \in \mathbb{R}^{T \times d} \) that is tailored to the specific characteristics of each method. We assume no domain shifts for $\bx_{t}$. The details of the data generation process are provided in Appendix \ref{app:data_numerical_1}. 

During the training process, we set $m_j=1$ meaning that no adaptive expansion of meta-domains occurs. For each epoch, we record the average expert weight $w_{j}(\calX_{t})$ over all $T$ training points for the five experts, as shown in Figure \ref{fig:interpretable weight}. From the results, we observe that for time series generated by PCA, TCPD, NMF, and SDL, their corresponding expert exihibits the highest weight. This indicates that CICADA effectively selects the expert that best describes the features of the time-series data. Furthermore, since PCA, TCPD, NMF, and SDL primarily extract linear features of data, their corresponding experts exhibit certain interchangeability. As a result, in addition to the expert generating the data, the other linear-feature experts also receive small but nonzero weights, as observed in Figure \ref{fig:interpretable weight}(d) and (e).
In contrast, for time-series data generated using Kernel PCA, which captures nonlinear features, it is challenging for any single expert to fully capture the underlying data structure. Consequently, all five experts contribute to feature extraction and reconstruction, as shown in Figure \ref{fig:interpretable weight}(c).
\begin{figure}[t]
    \centering
    \includegraphics[width=\linewidth]{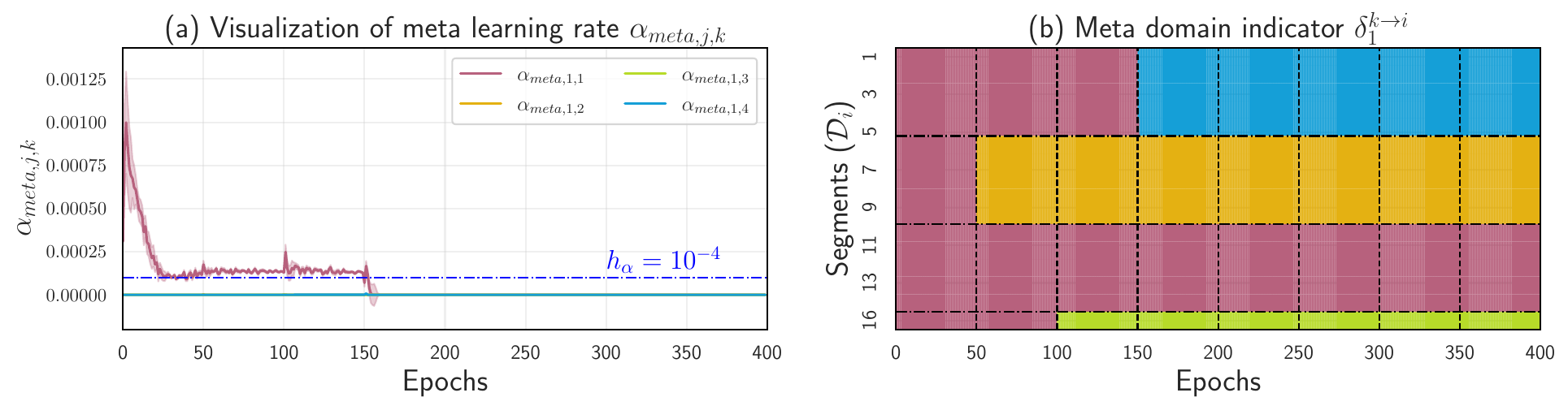}
    \vspace{-23pt}
    \caption{Meta-learning rate and meta-domain assignment for time-series generated by PCA.}
    \vspace{-14pt}
    \label{fig:adaptive pca}
\end{figure}

\begin{figure}[t]
    \centering
    \includegraphics[width=\linewidth]{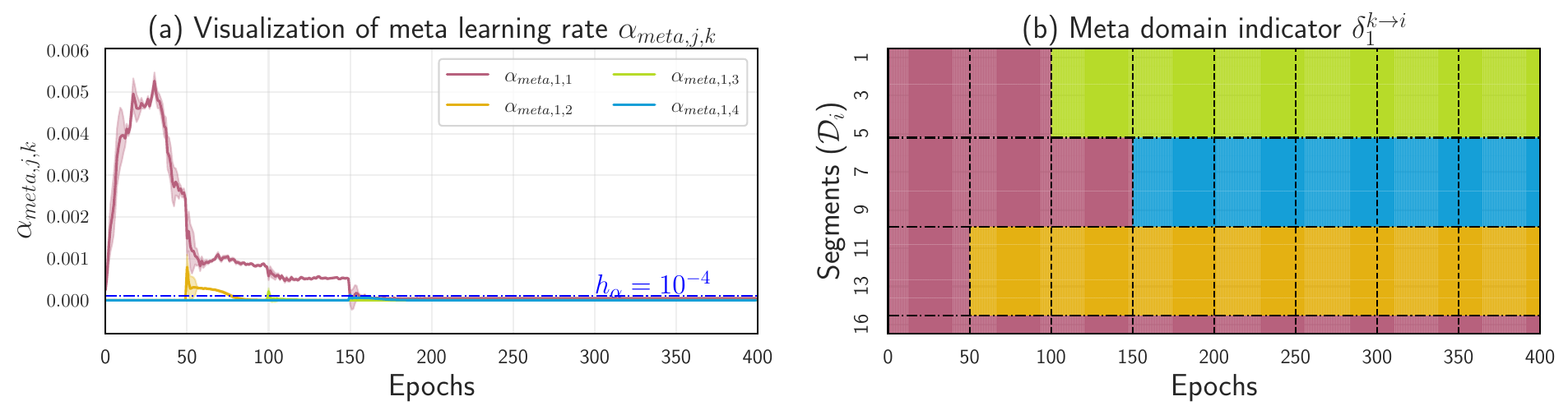}
    \vspace{-23pt}
    \caption{Meta-learning rate and meta-domain assignment for time-series generated by TCPD.}
    \vspace{-15pt}
    \label{fig:adaptive tensor}
\end{figure}

\subsection{Experiments for Selective Meta-Learning with Adaptive Expansion}
\label{sec:numerical_2}
The objective of this experiment is to demonstrate the efficiency of our selective meta-learning framework with adaptive expansion. For the time series generated by PCA and TCPD, we now consider \( \bX \) contain multiple domains. Specifically, $\calX$ contain 16 segments. The segments \( \calD_1 \) through \( \calD_5 \) belong to the first domain \( \bbP_1 \), \( \calD_6 \) through \( \calD_{10} \) belong to the second domain \( \bbP_2 \), \( \calD_{11} \) through \( \calD_{15} \) belong to the third domain \( \bbP_3 \), and \( \calD_{16} \) belongs to the fourth domain \( \bbP_4 \). 
The details of the data generation process are provided in Appendix \ref{app:data_numerical_2}. We begin with a single meta-domain $m=1$ and \( \Theta_1 = \{\Theta_1^1\} \), and allow CICADA to adaptive add meta-domains. The meta-learning rate is evaluated every 50 epochs to check if it exceeds the threshold. 

Figures \ref{fig:adaptive pca}(a) and \ref{fig:adaptive tensor}(a) show the meta-learning rate across epochs. Initially, \( m_1 = 1 \), and as shown in Figure \ref{fig:meta-learning}(c), the meta-learning rate \( \alpha_{\text{meta},1,1} \) is relatively high. This results in adaptive expansions occurring at the 50th, 100th, and 150th epochs, leading to \( m_j = 4 \) by the end of the 150th epoch. As the meta-learning rate drops rapidly afterward and approaches zero, it indicates that the expert has successfully learned sufficient meta-domains. In this case, because the four domains generating the time series are quite distinct, each domain is assigned to a separate meta-domain.

For a clearer illustration, Figures \ref{fig:adaptive pca}(b) and \ref{fig:adaptive tensor}(b) show the value of $\delta_{j}^{k \to i}$ of each $\calD_{i}$ over epochs. Initially, all the segments are assigned to the first meta-domain (red). As the second meta-domain (orange) is added at the 50th epoch, segments $\calD_{i}, \forall i = 11,\ldots 15$ are shifted to the second meta-domain. Then the third meta-domain (green) is introduced at the 100th epoch and includes $\calD_{i}, \forall i = 1,\ldots 5$. Finally, the fourth meta-domain (blue) is added at the 150th epoch and includes $\calD_{i}, \forall i = 6,\ldots 10$. This shows how the expert gradually learns the correct division of the bounded set $\calP$. 

\section{CASE STUDIES}
\label{sec:case}

\begin{figure}[t]
    \centering
    \includegraphics[width=\linewidth]{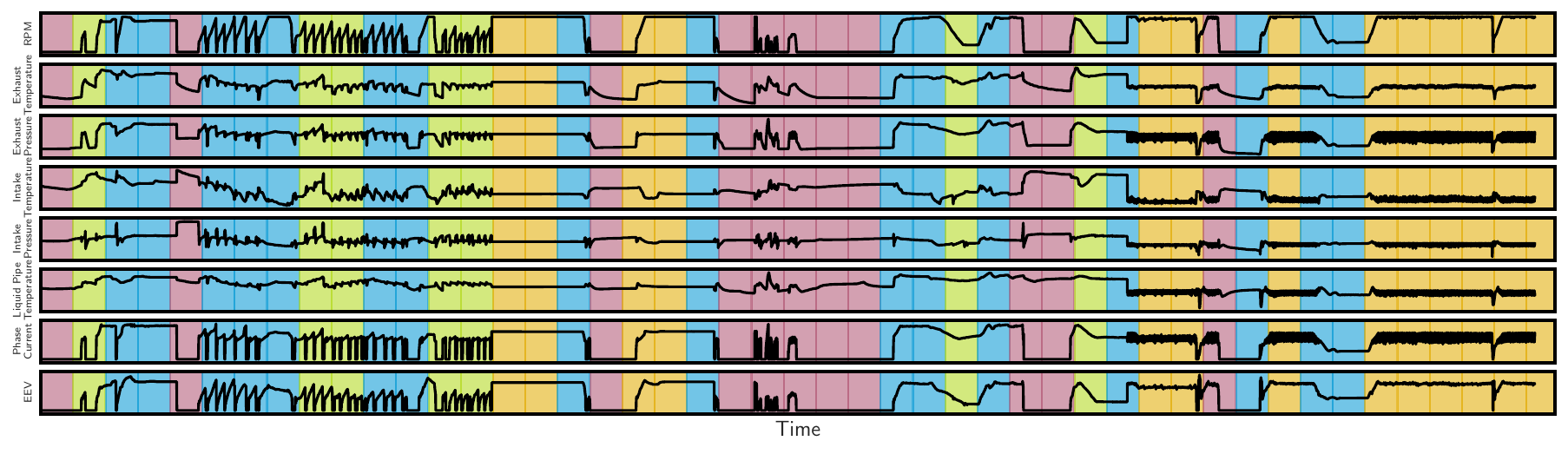}
    \vspace{-24pt}
    \caption{Different meta-domains detected by expert PCA for the Compressor Dataset}
    \vspace{-14pt}
    \label{fig:compressor-add pca}
\end{figure}

\begin{figure}[t]
    \centering
    \includegraphics[width=\linewidth]{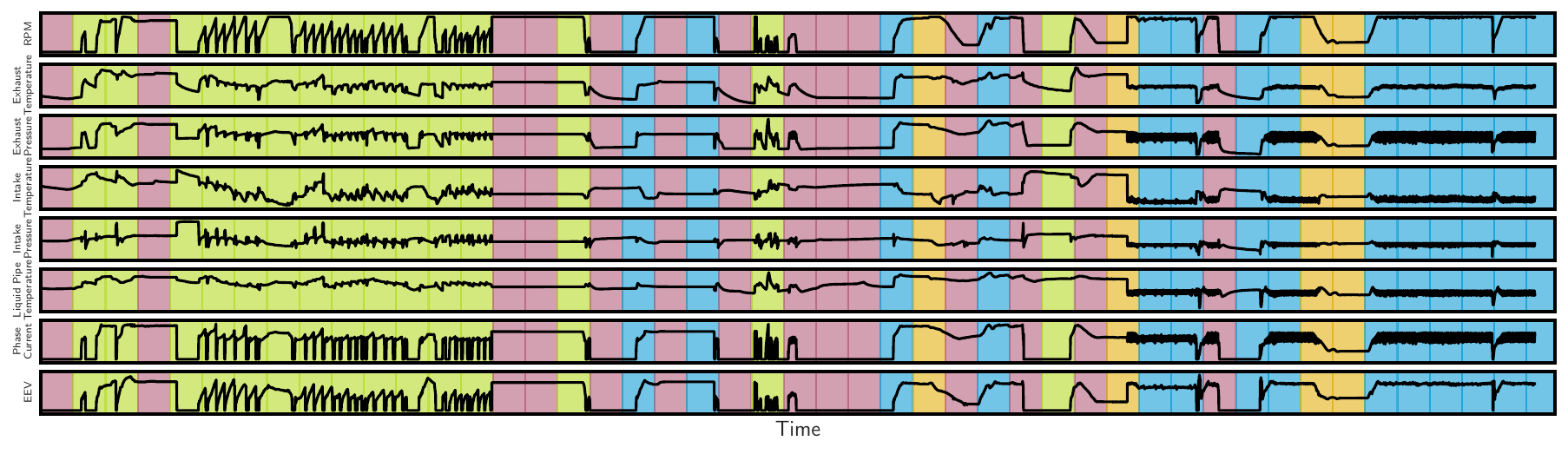}
    \vspace{-24pt}
    \caption{Different meta-domains detected by expert SFA for the Compressor Dataset}
    \vspace{-16pt}
    \label{fig:compressor-add sfa}
\end{figure}

\begin{figure*}[t]
    \centering
    \includegraphics[width=0.97\linewidth]{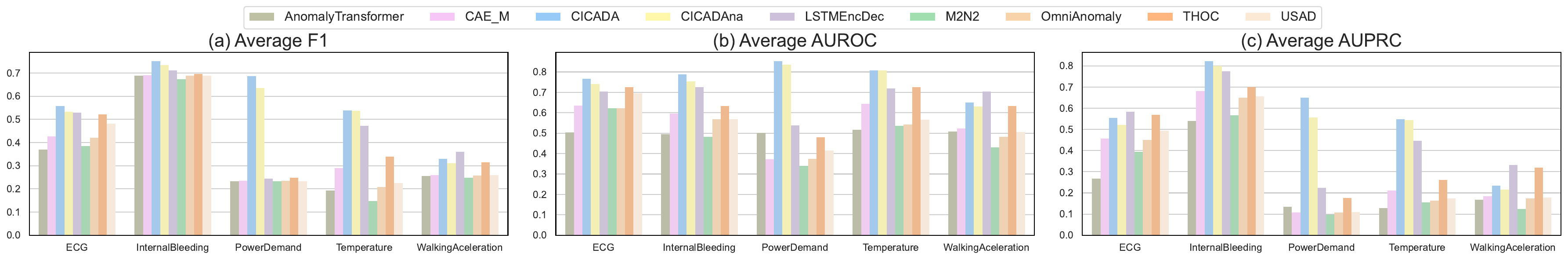}
    \vspace{-10pt}
    \caption{F1 score, AUROC and AUPRC for UCR Datasets}
    \vspace{-10pt}
    \label{fig:case1}
\end{figure*}

\begin{figure*}[t]
    \centering
    \includegraphics[width=0.97\linewidth]{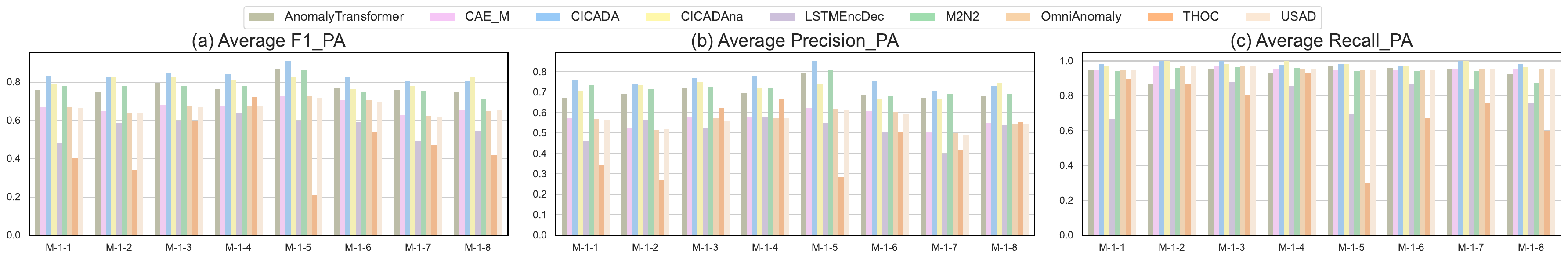}
    \vspace{-10pt}
    \caption{F1 score, precision and recall for SMD Datasets}
    \vspace{-13pt}
    \label{fig:case2}
\end{figure*}

\subsection{Experiments on Real Industrial Datasets}
We apply CICADA to three industrial manufacturing processes, each characterized by different operational scenarios that switch over time: 1) the FluorinePump Dataset, 2) the Compressor Dataset and 3) the Busbar Dataset. CICADA is implemented with experts including MLP, PCA, Kernel PCA, and Slow Feature Analysis (SFA). Due to space constraints, we focus on presenting the results for the Compressor Dataset, with the results and analysis for the other two datasets available in Appendix \ref{app:domain-division}.

\textbf{Detection of different meta-domains:} Figure \ref{fig:compressor-add pca} illustrates the four distinct meta-domains detected by expert PCA, which highlights different types of linear correlations between variables. These meta-domains can be interpreted in terms of the variable "RPM"(the first row) and correspond to different operational conditions. Specifically, the red meta-domain represents the sleep condition, where RPM is near zero. The yellow meta-domain corresponds to the stable operating condition, where RPM is high. The green and blue meta-domains indicate transition states, either from sleep to operating or from operating to sleep.

Figure \ref{fig:compressor-add sfa} shows the four meta-domains detected by the SFA expert, which offer a different perspective compared to PCA, as SFA focuses on the slowly evolving features of the time series. These four meta-domains can be interpreted in terms of the "speed of change" of the system. The green meta-domain corresponds to a phase of rapid change, while the blue meta-domain reflects slower changes. The red meta-domain indicates minimal change, and the yellow meta-domain represents a transitional state.

By integrating the meta-domains extracted from both experts, CICADA enables a more interpretable division of the domains.

\textbf{Anomaly detection performance:} We evaluate the anomaly detection performance for CICADA and compare it with the following state-of-the-art methods including prediction-based method: AnomalyTransformer~\citep{xu2021anomaly}, reconstruction-based methods: M2N2~\citep{kim2024model} (which employs test-time adaptation), LSTM~\citep{malhotra2016lstm}, OmniAnomaly~\citep{su2019robust}, USAD~\citep{audibert2020usad}, clustering-based method THOC~\citep{shen2020timeseries}, composite method: CAE-M~\citep{zhang2021unsupervised} and an ablation study CICADAna (not adding) that removes the adaptive expansion mechanism of CICADA. The results of F1 score, AUROC and AUPRC whose definitions are shown in Appendix \ref{app:metric} are shown in Table \ref{tab:industrial_anomaly_detection}. 

\begin{table}[t]
\caption{Anomaly detection performance for the three industrial datasets.}
\vspace{-10pt}
\centering
\scalebox{0.65}{
\begin{tabular}{lccccccccc}
\toprule
Baselines & \multicolumn{3}{c}{FluorinePump} & \multicolumn{3}{c}{Compressor} & \multicolumn{3}{c}{Busbar} \\
\cmidrule(lr){2-4} \cmidrule(lr){5-7} \cmidrule(lr){8-10}
 & F1 & AUROC & AUPRC & F1 & AUROC & AUPRC & F1 & AUROC & AUPRC \\
\midrule
M2N2 &  47.9  & 47.1 &  21.8       &  26.1   &   47.3 & 13.9  & 31.1 & \textbf{85.7} & 22.3\\
AT &  76.8  & 90.7 &  60.5       &  26.0  &   50.3 & 15.5  & 10.7 & 46.4 & 4.1 \\
THOC & 95.5 & 97.1 & 78.2    &  42.4    & 67.3  & 19.5  & 26.2 & 67.9 & 13.3 \\
USAD & 74.2 &  79.0 &  40.0     &   42.1   & 65.9  & 18.5  & 19.8 & 67.3 & 11.3 \\
LSTM & 67.1 &  84.7  &  62.8   &   \underline{52.5}   & \textbf{87.3}  & \textbf{58.2} & 30.2 & 74.3 & 22.9 \\
Omni & 50.9 &  45.7  &  21.7   &  40.6    & 60.6  &  16.3      & 20.8 & 71.1 & 11.5 \\
CAE-M &  53.9  &  49.1 &  22.9 &   40.7  &   56.6     & 15.3        & 20.0    & 69.8 & 11.7 \\
\midrule
CICADA & \textbf{99.91}  & \textbf{99.97}  & \textbf{99.92}        &   \textbf{65.6}  & \underline{80.6}       &  \underline{45.4}      &  \textbf{44.8}   &  \underline{77.6}      &  \textbf{31.1}    \\
CICADAna & \underline{99.4}  & \underline{99.7}  & \underline{96.3} & 36.5  & 63.8   &  25.3  &  \underline{39.3}   &  69.8    &  \underline{27.9}    \\
\bottomrule
\end{tabular}
}
\vspace{-15pt}
\label{tab:industrial_anomaly_detection}
\end{table}

CICADA significantly outperforms all the other baselines across all the three datasets, and achieves the highest F1 scores consistently, with particularly outstanding performance on the FluorinePump dataset. 
Compared to the methods without domain adaptation, CICADAna exhibits high robustness only by our MOE structure. Furthermore, CICADA demonstrates a notable improvement over CICADAna, highlighting the benefits of incorporating adaptive expansion. Although M2N2, which employs a test-time adaption strategy, can be used for domain adaptive and achieves good results on the Busbar dataset, its time-time adaption may be too flexible to adapt anomalous observations to normal ones, leading to high miss-detection rates. 

\subsection{Transferability Experiments on Public Datasets}
We further evaluate the effectiveness of CICADA using several public datasets, applying the same hyperparameters across all experiments. The experts used are PCA, Kernel PCA, and SFA, with parameters set as \(\lambda_1 = 1000\), \(\alpha_1 = 10^{-3}\),  \(h_{\alpha} = 5 \times 10^{-4}\), $L=10$ for the UCR datasets and $L=30$ for the SMD datasets.

\textbf{Univariate time series datasets:} The UCR Datasets 
consist of many univariate time series datasets from various domains such as biology, physics, and industrial systems \cite{dau2019ucr, keogh2021multi}. We select several datasets that share similar meanings but differ in their training sets, training on one dataset and testing on the others. Detailed information can be found in Appendix \ref{app:UCR}. For each method, we use the threshold that maximizes the F1 score to identify anomalies in testing datasets. The results including F1 score, AUROC, and AUPRC are shown in Figure \ref{fig:case1}. Across all the datasets, CICADA exhibits robust generalization performance, achieving high F1 scores.

\textbf{Multivariate time series datasets:} The SMD datasets consist of multivariate time-series data collected from 28 server machines in a data center \cite{su2019robust}. We use the datasets from Machine-1-1 (M-1-1) to Machine-1-8 (M-1-8),  training on one dataset and testing on another. Following the previous studies on this dataset~\cite{xu2021anomaly,yang2023dcdetector}, we evaluate the model performance using point-adjusted results, i.e., F1\_PA, Precision\_PA, Recall\_PA as defined in Appendix \ref{app:metric}. For each method, we use the 99.5th percentile of its training data as the threshold for identifying anomalies. The results are shown in Figure \ref{fig:case2}. As with earlier experiments, CICADA consistently outperforms all other methods, achieving the highest anomaly detection performance across all test cases.

\section{CONCLUSION}
\label{sec:conclusion}
This paper proposes CICADA, a framework for cross-domain anomaly detection in unsupervised multivariate time series, where data distributions shift dynamically across unknown domains. CICADA’s core innovations are as follows: (1) a mixture-of-experts (MOE) framework that flexibly and generally captures anomaly features across domains, while providing strong interpretability; (2) a selective meta-learning mechanism that mitigates negative transfer between dissimilar domains; (3) an adaptive expansion mechanism that automatically detects new domains; and (4) a hierarchical attention structure for quantifying expert contributions during fusion, further enhancing interpretability. 

Future work can explore two main directions: (1) extending the current framework to online learning scenarios with streaming data and (2) developing the weighting assignment algorithm based on anomaly injection and classification-based methods.

\clearpage
\bibliographystyle{ACM-Reference-Format}
\balance
\bibliography{main}

\newpage

\appendix
\onecolumn
\noindent \textbf{\Large APPENDIX}
\section{Algorithms for CICADA}
\label{app:algorithms}
\vspace{-10pt}
\begin{algorithm}[H]
\caption{Training Process for CICADA}
\label{alg:CICADA}
\begin{algorithmic}[1]
\REQUIRE Multivariate time-series training data \(\mathbf{X}\), window size \(L\), maximum training epoch max\_epoch, adaptive expansion epoch epoch\_add, expansion threshold $h_{\alpha}$.
\STATE Generate \(\mathcal{X}_t\) by \(\mathcal{X}_t = (\mathbf{x}_t, \mathbf{x}_{t-1}, \ldots, \mathbf{x}_{t-L})^\top\).
\STATE Partition segments \(\mathcal{D}_1, \mathcal{D}_2, \ldots, \mathcal{D}_N\) from $\calX$.
\FOR{epoch = 1 to max\_epoch}
    \STATE Sample \(\mathcal{D}_i\) from \(\mathcal{D}_1, \mathcal{D}_2, \ldots, \mathcal{D}_N\).
    \STATE Generate \(\mathcal{D}_i^{\text{tr}}\) and \(\mathcal{D}_i^{\text{val}}\) from \(\mathcal{D}_i\).
    \STATE Sample \(\mathcal{X}_t\) from \(\mathcal{D}_i^{\text{val}}\).
    \FOR{$j = 1$ to \(m\)}
        \STATE Obtain \(\{\Theta_j^{1\rightarrow i},\Theta_j^{2\rightarrow i},\ldots,\Theta_j^{m_j\rightarrow i}\}, \mathcal{L}_{\text{meta}, j}\) by Algorithm \ref{alg:selective meta-learning}.
        \FOR{$k=1$ to $m_j$}
        \STATE Compute extracted features for expert $j$ by $\mathbf{f}_{j,t}^k=g_j(\calX_t;\Theta_j^{k \rightarrow i})$.
        \ENDFOR
        \STATE Compute $\bz_{j,t}=\text{Multihead-Attention}(\calX_t,\mathbf{f}_{j,t},\mathbf{f}_{j,t})$.
    \ENDFOR
    \STATE Compute \(\tilde{\mathbf{z}}_t = \text{Multihead-Attention}(\mathcal{X}_t, \mathbf{z}_t, \mathbf{z}_t)\)
    \STATE Compute \(\mathcal{L}_{\text{reconstruction}}\) and \(\mathcal{L}_{\text{extraction}}\).
    \STATE Update parameters $\Xi_{j}$, $\boldsymbol{\alpha}_j$, $\bW^Q_{l,j}$, $\bW^K_{l,j}$, $\bW^V_{l,j}$ for $j=1,\ldots, m, l = 1,\ldots, h$, $\tilde{\bW}^Q_l$, $\tilde{\bW}^K_{l}$ and $\tilde{\bW}^V_l$ for $l = 1,\ldots, \tilde{h}$, $\Theta_{\text{MLP}}$ by backward propagation.
    \IF{epoch \% epoch\_add = 0}
        \FOR{$j = 1$ to \(m\)}
            \STATE \(k = \underset{k}{\arg \max} \ \alpha_{\text{meta}, j, k}\).
            \IF{\(\alpha_{\text{meta},j,k} > h_\alpha\)}
                \STATE Find \(i_{\text{farthest}}\) by $i_{\text{farthest}}=\underset{\{i=1,\ldots, N|\delta_{j}^{k\to i}=1\}}{\arg \max} \ \| \Theta_j^{k \rightarrow i} - \Theta_j^k \|_F^2$.
                \STATE Generate \(\Theta_j^{\text{new}} = \Theta_j^{k \rightarrow i_{\text{farthest}}}\).
                \STATE \(\Xi_j = \Xi_j \cup \{ \Theta_j^{\text{new}} \}\).
                \STATE \(m_j = m_j + 1\).
            \ENDIF
        \ENDFOR
    \ENDIF
\ENDFOR
\end{algorithmic}
\end{algorithm}
\vspace{-10pt}

\begin{algorithm}[H]
\caption{Selective Meta-Learning Algorithm}
\label{alg:selective meta-learning}
\begin{algorithmic}[1]
\REQUIRE The meta-training set $\mathcal{D}_i^{\text{tr}}$, and the meta-validation set $\mathcal{D}_i^{\text{val}}$, the number of meta-domains $m_j$ for $j=1,\ldots, m$. 
\STATE Compute the loss for each meta-domain: \(\mathcal{L}_j(\mathcal{D}_i^{\text{tr}}; \Theta_j^k)\) for all \(k = 1\) to \(m_j\).
\STATE Compute the transfer indicator \(\delta_{j}^{k \to i}\) based on \(\mathcal{L}_j(\mathcal{D}_i^{\text{tr}}; \Theta_j^k), k = 1,\ldots, m_j\).
\FOR{$k = 1$ to \(m_j\)}
    \IF{$\delta_j^{k \rightarrow i} = 1$}
        \STATE \(\Theta_j^{k \rightarrow i} = \Theta_j^k - \alpha_{{\rm meta}, j, k} \nabla_{\Theta_j^k} \mathcal{L}_j(\mathcal{D}_i^{\rm tr}; \Theta_j^k)\).
    \ELSE
        \STATE \(\Theta_j^{k \rightarrow i} = \Theta_j^k\).
    \ENDIF
\ENDFOR
\STATE Compute $\mathcal{L}_{\text{meta}, j}(\mathcal{D}_j^{\text{val}}; \Xi_j)$ based on $\Theta_j^{k \rightarrow i},k=1,2,\ldots,m_j$.
\RETURN $\mathcal{L}_{\text{meta}, j}(\mathcal{D}_j^{\text{val}}; \Xi_j)$.
\end{algorithmic}
\end{algorithm}

\section{Gradient-based Statistical Experts and Their Anomaly Score Functions}
\label{app:anomaly_score_function}

Appendices \ref{app:pca} to \ref{app:sdl} introduce six commonly used machine learning-based experts, and Appendix \ref{app:deep-learning} discusses how to add the deep learning-based experts to CICADA.

\subsection{PCA}
\label{app:pca}
In Principal Component Analysis (PCA), the objective is to solve the following optimization problem:  
\begin{equation*}
\begin{split}
    \max_{\bW} \ {\rm tr}(\bW^\top \bSigma \bW), \\
    \text{subject to} \quad  \bW^\top \bW = \bI,    
\end{split}
\end{equation*}
where \(\bSigma\) is the covariance matrix of the data. To address this problem, we employ the projected gradient descent algorithm. By defining the loss function \(\calL_j\) as:
\begin{equation*}
    \calL_j = - {\rm tr}({\rm QR}(\bW)^\top {\rm vec}(\calX_t) {\rm vec}(\calX_t)^\top {\rm QR}(\bW)),
\end{equation*}
we directly train PCA using a gradient-based algorithm, where {\rm QR}($\cdot$) refers to the QR decomposition.

The feature \(\mathbf{f}_{j,t}\) can be represented by:
\[
\mathbf{f}_{j,t} = {\rm Linear}({\rm vec}(\calX_t) {\rm QR}(\bW)),
\]
where ${\rm Linear}(\cdot)$ represents the linear transform to project the score to the $K$-dimension space.

Its anomaly score function is given by:
\begin{equation*}
    \text{anosc}(\calX_t) = \| {\rm vec}(\calX_t) - {\rm vec}(\calX_t) {\rm QR}(\bW) {\rm QR}(\bW)^\top \|_2^2.
\end{equation*}

\subsection{Kernel PCA}
In Kernel PCA, the objective is to maximize the variance in a higher-dimensional feature space using a kernel function. The optimization problem is:
\begin{equation*}
\begin{split}
    \max_{\bW} \ {\rm tr}(\bW^\top \bK \bW), \\
    \quad \text{subject to} \quad \bW^\top \bW = \bI,    
\end{split}
\end{equation*}
where \(\bK\) is the kernel matrix, computed as \(\bK = \Phi(\calX) \Phi(\calX)^\top\), with \(\Phi(\calX)\) mapping the input data to a higher-dimensional feature space. To solve this problem, we use the Kernel Hebbian Algorithm given in ~\cite{kim2003kernel}.

The feature \(\mathbf{f}_{j,t}\) can be expressed as:
\[
\mathbf{f}_{j,t} = {\rm Linear}(\bK \bW),
\]
where ${\rm Linear}(\cdot)$ represents linear transform to project the score to the $K$-dimension space.

The anomaly score function for Kernel PCA is given by:
\begin{equation*}
    \text{anosc}(\calX_t) = \| {\rm vec}(\calX_t) - \text{MLP}_j(\mathbf{f}_{j,t}) \|_2^2.
\end{equation*}

\subsection{SFA}
In Slow Feature Analysis (SFA), the objective is to find a linear transform of the data that minimizes the change rate over time. The optimization problem can be framed as:
\begin{equation*}
\begin{split}
    \min_{\bW} \ \left[ \bW^\top (\bx_t - \bx_{t-1}) \right], \\
    \quad \text{subject to} \quad \bW^\top \bW = \bI,    
\end{split}
\end{equation*}
where \(\bx_t\) represents the data at time \(t\). The loss function \(\calL_j\) can be written as:
\begin{equation*}
    \calL_j = {\rm tr}({\rm QR}(\bW)^\top \Delta \bx_t \Delta \bx_t^\top {\rm QR}(\bW)),
\end{equation*}
where \(\Delta \bx_t = \bx_t - \bx_{t-1}\) denotes the change in the data over time.

The feature \(\mathbf{f}_{j,t}\) can be represented by:
\[
\mathbf{f}_{j,t} = {\rm Linear}(\Delta \bx_t {\rm QR}(\bW)),
\]
where ${\rm Linear}(\cdot)$ represents linear transform to project the score to the $K$-dimension space.

The anomaly score function for SFA is defined as the $T^2$ statistic of feature $\mathbf{f}_{j,t}$:
\begin{equation*}
    \text{anosc}(\calX_t) = (\mathbf{f}_{j,t}-\bar{\mathbf{f}}_j)^\top \bSigma_{\mathbf{f}} (\mathbf{f}_{j,t}-\bar{\mathbf{f}}_j),
\end{equation*}
where $\bar{\mathbf{f}}_{j}$ is the mean vector of $\mathbf{f}_{j,t}$ and $\bSigma_{\mathbf{f}}$ is the covariance matrix of $\mathbf{f}_{j,t}$, which can be recorded during the training process.

\subsection{NMF}
In Non-negative Matrix Factorization (NMF), the goal is to decompose the input matrix \(\calX_t\) into two non-negative matrices \(\bW\) and \(\bH\) such that:
\begin{equation*}
    {\rm vec}(\calX_t) \approx \bW \bH.
\end{equation*}
The optimization problem is:
\begin{equation*}
\begin{split}
    \min_{\bW, \bH} \| \calX_t - \bW \bH \|_F^2, \\
    \text{subject to} \quad \bH \geq 0,
\end{split}
\end{equation*}
where \(\| \cdot \|_F\) is the Frobenius norm, and the condition \(\bH \geq 0\) implies element-wise non-negativity.

The loss function \(\calL_j\) can be represented as:
\begin{equation*}
    \calL_j = \| {\rm vec}(\calX_t) - {\rm vec}(\calX_t) {\rm ReLU}(\bH)^\dagger {\rm ReLU}(\bH) \|_F^2,
\end{equation*}
where \({\rm ReLU}(\cdot)\) refers to the element-wise application of the ReLU activation function. For a matrix $\bH$, $\bH^\dagger=(\bH^\top\bH)^{-1}\bH^\top$.

The feature \(\mathbf{f}_{j,t}\) can be expressed by:
\[
\mathbf{f}_{j,t} = {\rm Linear}({\rm vec}(\calX_t) {\rm ReLU}(\bH)^\dagger),
\]
where ${\rm Linear}(\cdot)$ represents linear transform to project the score to $K$-dimension space.

The anomaly score function for NMF is given by:
\begin{equation*}
    \text{anosc}(\calX_t) = \| {\rm vec}(\calX_t) - {\rm vec}(\calX_t) {\rm ReLU}(\bH)^\dagger {\rm ReLU}(\bH) \|_F^2.
\end{equation*}

\subsection{TCPD}
In Tensor CP Decomposition (TCPD), the objective is to factorize a tensor into a sum of component rank-1 tensors. Specifically, we aim to solve the following optimization problem:
\[
    \min_{\bA, \bB} \| \calX_t - \sum_{r=1}^R \lambda_i \cdot\mathbf{a}_r \circ \mathbf{b}_r  \|_F^2,
\]
where \(\calX_t\) is the input tensor, and \(\mathbf{a}_r, \mathbf{b}_r\) are the factors corresponding to the three modes of the tensor, with \(\circ\) denoting the outer product. We assume $\|\ba_r\|_2=\|\bb_r\|_2=1,\forall r=1,2,\ldots,R$. The goal is to find the factors \(\bA, 
\bB\) such that their outer product closely approximates the input tensor. 

The loss function $\calL_j$ can be written as:
\[
    \calL_j = \|{\rm vec}(\calX_t) - {\rm vec}(\calX_t) \bW^\dagger \bW \|_F^2,
\]
\[
\bW={\rm Reshape}\left(\left[\frac{\ba_r}{\|\ba_r\|_2} \circ \frac{\bb_r}{\| \bb_r\|_2}\right]\right)\in \bbR^{Ld\times R}.
\]

The feature \(\mathbf{f}_{j,t}\) can be expressed by:
\[
    \mathbf{f}_{j,t}= {\rm Linear}({\rm vec}(\calX_t) \bW^\dagger),
\]
where ${\rm Linear}(\cdot)$ represents linear transform to project the score to the $K$-dimension space.

The anomaly score function for TCPD is given by:
\[
    \text{anosc}(\calX_t) = \|{\rm vec}(\calX_t) - {\rm vec}(\calX_t) \bW^\dagger \bW \|_F^2.
\]

\subsection{SDL}
\label{app:sdl}
In Sparse Dictionary Learning (SDL), the goal is to represent the data matrix \(\calX_t\) as a sparse linear combination of a dictionary matrix \(\bD\) and a sparse coefficient matrix \(\bH\). The optimization problem is formulated as:
\[
\begin{split}
    \min_{\bD, \bH} \| \calX_t - \bD \bH \|_F^2 \\ \quad \text{subject to} \quad \| \bH \|_0 \leq k,
\end{split}
\]
where \(\| \bH \|_0\) denotes the number of non-zero entries in the coefficient matrix \(\bH\), and \(k\) is a predefined sparsity level. The goal is to find the dictionary \(\bD\) and the sparse codes \(\bH\) that best reconstruct \(\calX_t\), while ensuring that the coefficients \(\bH\) remain sparse.

By Lagrange dual and $l_1$ approximation, we construct the loss function by:
\[
\calL_j = \| {\rm vec}(\calX_t) - {\rm vec}(\calX_t) \bH^\dagger \bH\|_F^2 + \lambda_{SDL} \| \bH\|_1,
\]
where $\lambda_{SDL}$ is the regularization parameter to ensure the sparsity of $\bH$. 

The feature \(\mathbf{f}_{j,t}\) can be expressed by:
\[
    \mathbf{f}_{j,t}= {\rm Linear}({\rm vec}(\calX_t) \bH^\dagger),
\]
where ${\rm Linear}(\cdot)$ represents linear transform to project the score to the $K$-dimension space.

The anomaly score function for SDL is given by:
\[
    \text{anosc}(\calX_t) = \| {\rm vec}(\calX_t) - {\rm vec}(\calX_t) \bH^\dagger \bH\|_F^2,
\]
where \(\hat{\bH}\) is the sparse coefficient matrix corresponding to the test input \(\calX_t\).

\subsection{Deep Learning Models}
\label{app:deep-learning}
We begin by discussing how deep neural networks construct the extraction loss function, using the encoder as an example. For the encoder, the loss function is defined as
\[
\mathcal{L}_j = \| \text{vec}(\mathcal{X}_t) - \text{Decoder}(\mathbf{f}_{j,t}) \|_2^2,
\]
where \( \mathbf{f}_{j,t} = g_j(\text{vec}(\mathcal{X}_t)) \), and \( g_j \) represents an encoder. In order to optimize the extraction loss, we need to train a corresponding decoder to reconstruct the data from \( \mathbf{f}_{j,t} \). Therefore, in the training process in Algorithm \ref{alg:CICADA}, for each \( \Theta_j^k \), we must train the corresponding parameters \( \Theta_j^{de(k)} \) in the decoder. If a new meta-domain with new parameters is added, the corresponding parameters in the decoder must also be added and trained. The training process can be directly performed by optimizing the extraction loss.

The anomaly score for the encoder is given by
\[
{\rm anosc}(\mathcal{X}_t) =  \| \text{vec}(\mathcal{X}_t) - \text{Decoder}(\mathbf{f}_{j,t}) \|_2^2.
\]

For other deep neural network models, since feature extraction-based models are commonly used for anomaly detection, they can be constructed in a similar manner. However, depending on the model type, the structure of the model and the extraction loss \( \mathcal{L}_j \) may have different forms.



\section{Data Generation Process for Numerical Studies}
\subsection{Synthetic Data for Interpretable Expert Weight} 
\label{app:data_numerical_1}

In this section, we introduce the data generation process used in Section \ref{sec:numerical_1}. The data generation methods for PCA, NMF, and SDL follow the same framework, as they all aim to extract linear features. Specifically, we first generate $p$ distinct latent time series $\bz_t \in \mathbb{R}^p$ using the following equation:

\begin{equation*}
\bz_{j,t}=\begin{cases}
    \sin \frac{100jt}{2T} + \cos \sin \frac{123t}{T} + 0.1 & j=1,3,5,\ldots\\
    \cos \frac{100jt}{2T} + \sin \cos \frac{131t}{T} + 0.1 & j=0,2,4,\ldots
\end{cases}    
\end{equation*}

Next, we derive $\bx_t \in \mathbb{R}^d$ through a linear transform of $\bz_t$:
\begin{equation*}
    \bx_t = \bC \bz_t + 0.1 \boldsymbol{\varepsilon}_t,
\end{equation*}
where $\boldsymbol{\varepsilon}_t$ is the noise at time $t$ following the standard multivariate normal distribution with mean $\mathbf{0}$ and covariance matrix $\mathbf{I}$.

For the experts of \textbf{PCA, NMF and SDL}, we generate $\bC \in \mathbb{R}^{d \times p}$ in the following ways:
\begin{itemize}
    \item \textbf{PCA:} each element of $\bC$ is generated from a normal distribution.
    \item \textbf{NMF:} each element of $\bC$ is generated from a uniform distribution.
    \item \textbf{SDL:} each element of $\bC$ is generated from a normal distribution, with each row retaining only three non-zero elements.
\end{itemize}

For \textbf{TCPD}, we generate each time block in a bilinear manner as follows:
\[
\bb_t = \mathrm{vec}(\bA \bz_t \mathbf{B}) + 0.1 \boldsymbol{\varepsilon}_t \in \mathbb{R}^L.
\]
Next, we concatenate the time blocks to obtain:
\[
\bX = (\bb_1^\top, \bb_2^\top, \ldots, \bb_T^\top)^\top \in \mathbb{R}^{LT}.
\]

For \textbf{Kernel PCA}, we directly generate the time-series data $\bx_t \in \bbR^d$ by:
\begin{equation*}
\bx_{j,t}=\begin{cases}
    \sin \frac{100jt}{2T} + \cos \sin \frac{123t}{T} + 0.1 & j=1,3,5,\ldots\\
    \cos \frac{100jt}{2T} + \sin \cos \frac{131t}{T} + 0.1 & j=0,2,4,\ldots
\end{cases}    
\end{equation*}
Finally, we conclude the configuration for different data generation methods in Table \ref{tab:numercal_1-data}.
\begin{table}[h]
\centering
\caption{Configuration for Different Data Generation Methods}
\label{tab:numercal_1-data}
\begin{tabular}{ccccc}
\toprule
\textbf{Dataset}   & \textbf{Length} $T$ & \textbf{Window Size} $L$ & \textbf{Number of Latent Variables} $p$ & \textbf{Dimensions} $d$ \\ \midrule
\textbf{TCPD}   & 8000 & 10 & 5 & 40\\
\textbf{PCA}        & 800 & 1 & 5 & 40  \\
\textbf{KPCA}       & 800 & 1 & NA & 40 \\
\textbf{NMF}        & 800 & 1 & 10 & 40 \\
\textbf{SDL} & 800 & 1 & 10 & 40 \\ 
\bottomrule
\end{tabular}
\end{table}

\subsection{Synthetic Data for Selective Meta-Learning with Adaptive Expansion}
\label{app:data_numerical_2}
Section \ref{sec:numerical_2} focuses on synthetic data generated by PCA and Tensor CP decomposition. In this section, we insert four domains into $\bX$, with $\bbP_1$ to $\bbP_3$ each occupying 31.25\%, and $\bbP_4$ occupying 6.25\%. Within the different domains, the generation method for $\bz_{j,t}$ follows the procedure in Appendix \ref{app:data_numerical_1}, yet with different values of $\bA, \bB, \bC$. For domain $i$, the time-series data generated by PCA is given by:
\[
\bx_{t}= \bC_i \bz_t + \boldsymbol{\varepsilon}_t.
\]
Similarly, the time-series data generated by TCPD is given by:
\[
\bx_t = \mathrm{vec}(\bA_i \bz_t \bB_i) + 0.1 \boldsymbol{\varepsilon}_t \in \mathbb{R}^L.
\]

\section{Supplementary for Case Studies}
\subsection{Details for Performance Metrics}
\label{app:metric}
The definitions of the performance metrics used in Section \ref{sec:case} are as follows:
\begin{itemize}
    \item \textbf{F1 score}: The F1 score is the harmonic mean of Precision and Recall, providing a single metric to evaluate the balance between precision and recall. It is particularly useful when the classes are imbalanced. The formula for F1 score is:
    \[
    F1 = 2 \times \frac{\text{Precision} \times \text{Recall}}{\text{Precision} + \text{Recall}}.
    \]
    
    \item \textbf{AUROC (Area Under the Receiver Operating Characteristic Curve)}: AUROC is a metric that evaluates the ability of a classifier to distinguish between positive and negative classes. It plots the true positive rate (TPR) against the false positive rate (FPR) at various thresholds. The AUROC score ranges from 0 to 1, with a higher score indicating better model performance, with 1 being the ideal value.
    
    \item \textbf{AUPRC (Area Under the Precision-Recall Curve)}: AUPRC is the area under the curve that plots Precision against Recall at various thresholds. It is especially useful when dealing with imbalanced datasets, where the positive class is much less frequent than the negative class. A higher AUPRC score indicates better performance, with 1 being the ideal value.
    
    \item \textbf{F1\_PA, Precision\_PA and Recall\_PA}: The point adjustment method assumes that the detected anomaly time points do not need to fall within the correct anomaly time intervals precisely. Instead, the point adjustment method considers the detected anomaly time points that fall near to the true anomaly time intervals as correct detections. F1\_PA, Precision\_PA and Recall\_PA represent the F1 score, precision and recall values after applying the point adjustment method. 
\end{itemize}

\subsection{Details for Industrial Datasets}
\begin{table*}[h]
\centering
\caption{Experimental details for industrial datasets}
\label{tab:numercal_2-data}
\begin{tabular}{cccccc}
\toprule
\textbf{Dataset}   & \textbf{Training Length} $T$ & \textbf{Test length} $T'$ & \textbf{Window Size} $L$ & \textbf{Experts used} & \textbf{Dimensions} $d$ \\ \midrule
\textbf{FluorinePump}   & 16209 & 2366 & 2 & MLP, PCA, Kernel PCA & 9 \\
\textbf{Compressor}       & 29605 & 4051 & 5 & PCA, Kernel PCA, SFA & 8 \\
\textbf{Busbar}        & 6768 & 9750 & 10 & PCA, Kernel PCA, SFA & 15  \\
\bottomrule
\end{tabular}
\end{table*}

\subsection{Results for Meta-domain Detection in Industrial Datasets}
\begin{figure}[h]
    \centering
    \includegraphics[width=0.6\linewidth]{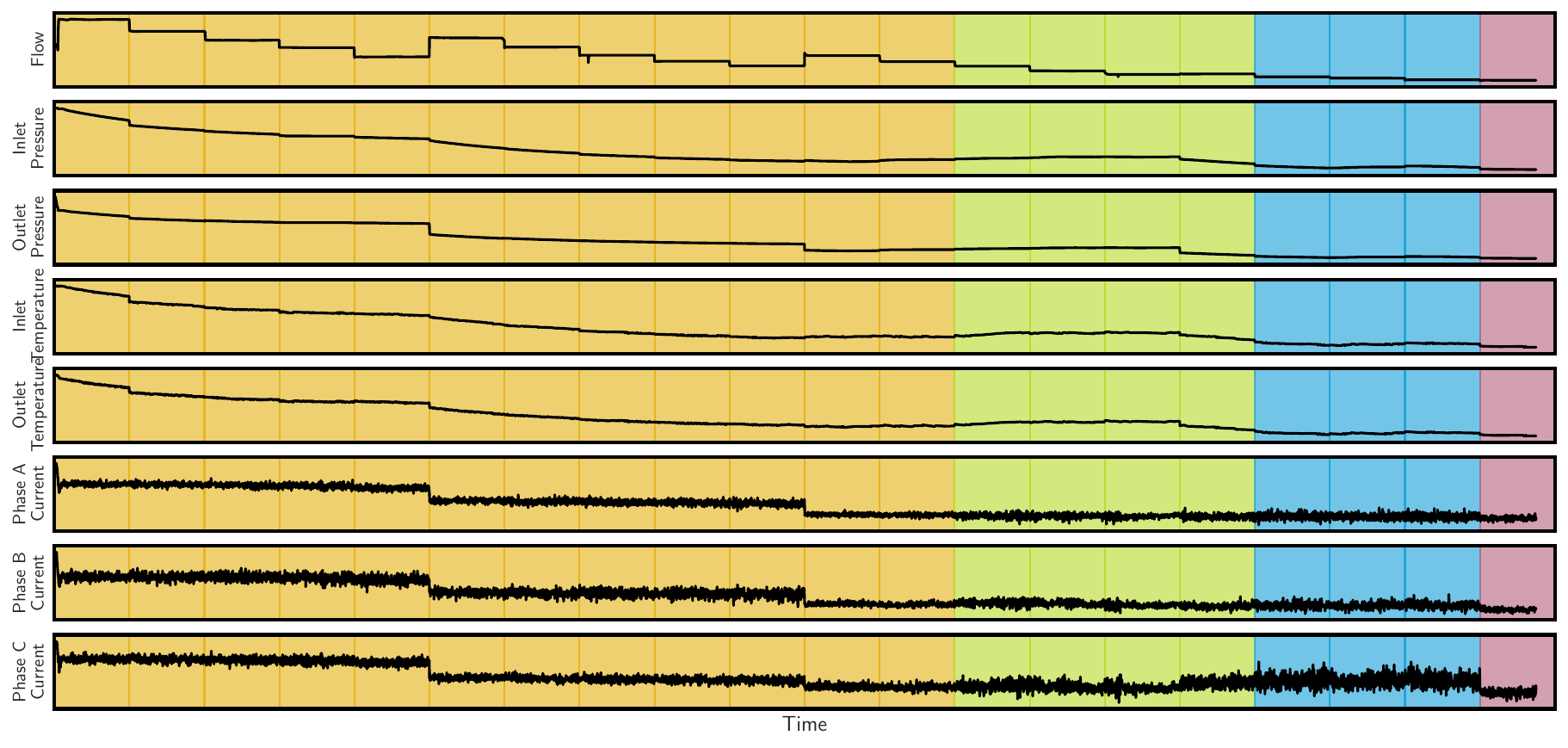}
    \vspace{-10pt}
    \caption{Different meta-domains detected by expert PCA for the FluorinePump Dataset}
    \vspace{-5pt}
    \label{fig:FluorinePump-add sfa}
\end{figure}

\begin{figure}[h]
    \centering
    \includegraphics[width=0.6\linewidth]{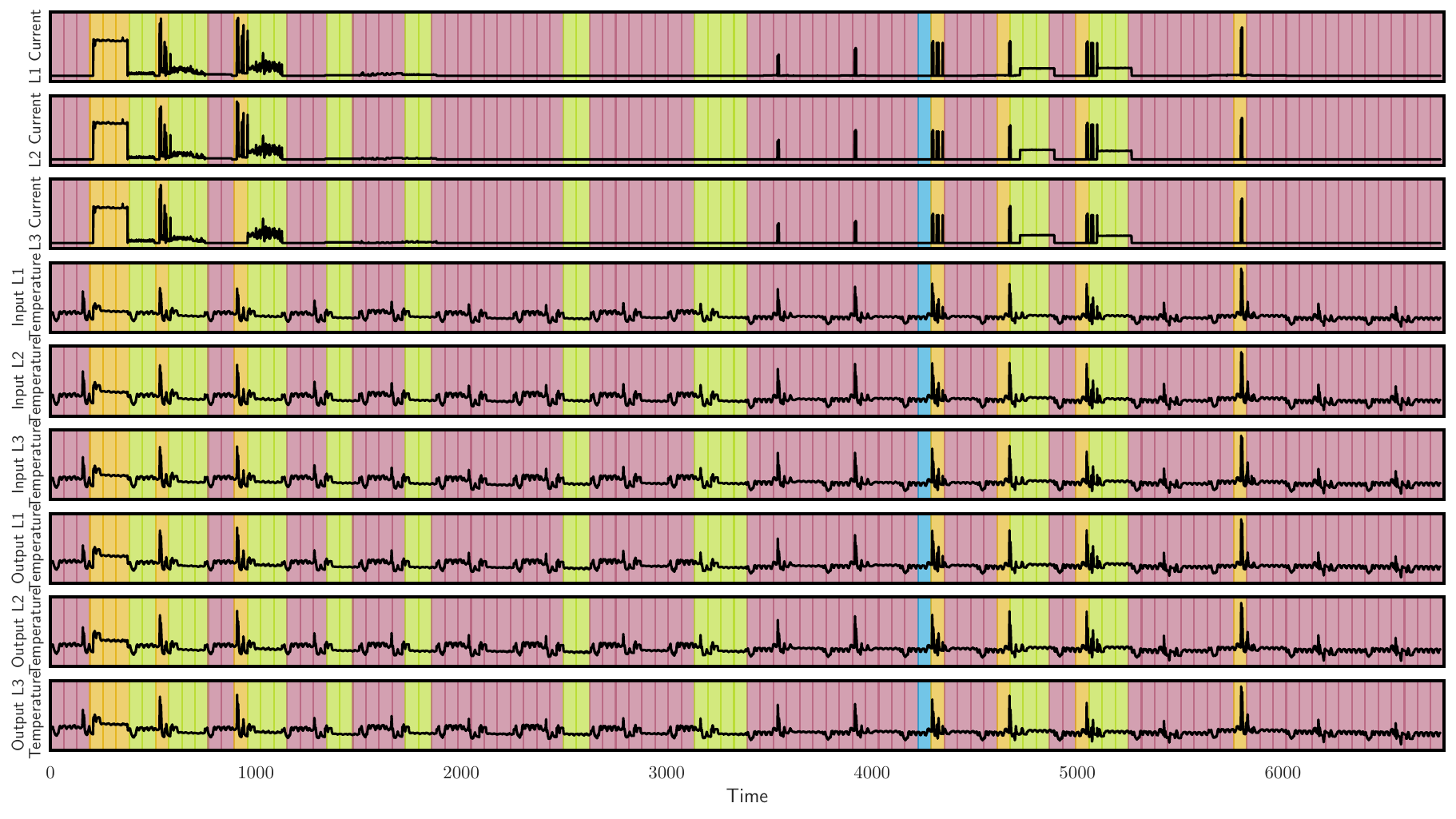}
    \vspace{-10pt}
    \caption{Different meta-domains detected by expert PCA for the Busbar Dataset}
    \vspace{-5pt}
    \label{fig:busbar-add pca}
\end{figure}

\begin{figure}[h]
    \centering
    \includegraphics[width=0.6\linewidth]{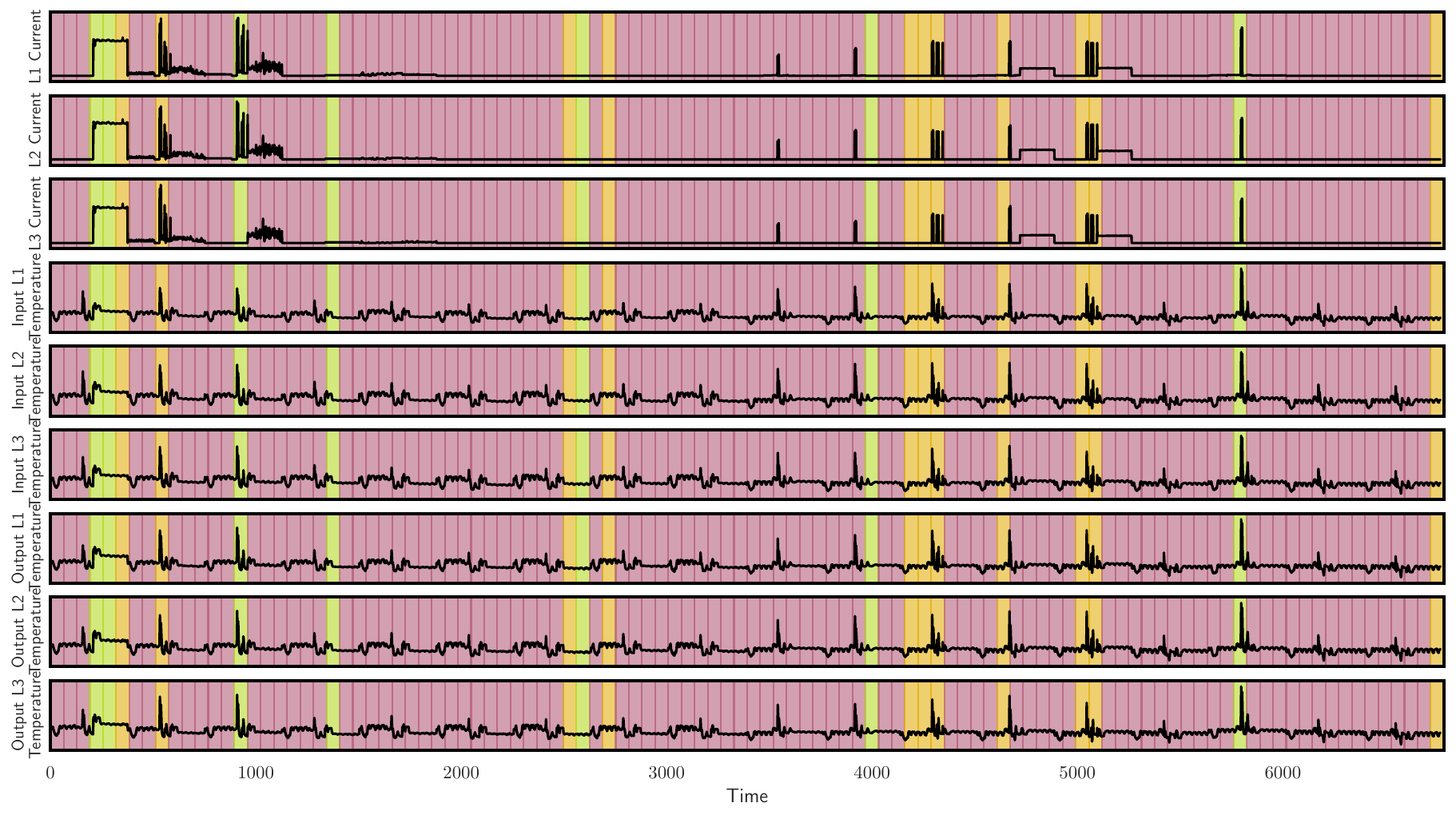}
    \vspace{-10pt}
    \caption{Different meta-domains detected by expert SFA for the Busbar Dataset}
    \vspace{-5pt}
    \label{fig:Busbar-add sfa}
\end{figure}

Figures \ref{fig:FluorinePump-add sfa} through \ref{fig:Busbar-add sfa} demonstrate comparative meta-domain division results of different algorithms for the industrial datasets by giving the first eight variables. 

Figure \ref{fig:FluorinePump-add sfa} presents the different meta-domains detected by PCA expert for the FluorinePump dataset. The temporal progression reveals systematic migration through distinct operational regimes. Since expect SFA only detects one meta-domain, we do not show its result here. This can be intuitively understood by the small data variations over time. 

Figures \ref{fig:busbar-add pca} to \ref{fig:Busbar-add sfa} present the different meta-domains detected by PCA expert and SFA expert for the Busbar Dataset. In Figure \ref{fig:busbar-add pca}, the ``current'' variables (including the first row to the third row) distinguish different electrical load conditions: red (low current), yellow (medium current), green (high current), and blue (transitional states). Figure \ref{fig:Busbar-add sfa} also exhibits distinct operational patterns by ``current'' variables, where red denotes stable operational conditions, while green and yellow correspond to differential transitional states. 

\label{app:domain-division}

\FloatBarrier

\subsection{Details for Datasets in the UCR Datasets}
\label{app:UCR}
    The experimental details for the datasets we use in the UCR datasets are shown in Table \ref{tab:case_1-data}.
\begin{table*}[h]
\centering
\caption{Experimental details for the UCR Datasets}
\label{tab:case_1-data}
\begin{tabular}{ccc}
\toprule
\textbf{Name}   & \textbf{Source Dataset ID} & \textbf{Target Dataset ID} \\ \midrule 
\textbf{ECG} & 32 & 33, 34, 35, 36 \\ 
\textbf{Internel Bleeding}   & 119 & 11, 12, 13, 14 \\
\textbf{Power Demand}       & 44 & 45, 46, 47\\
\textbf{Temperature}        & 113 & 5, 6, 7, 8, 9, 10 \\
\textbf{Walking Aceleration} & 53 & 54 \\
\bottomrule
\end{tabular}
\end{table*}
\section{Evolutionary Insight for Selective Meta-Learning and Adaptive Expansion}
\label{app:evolutionary-insight}
\begin{figure}[h]
    \centering
    \includegraphics[width=0.6\linewidth]{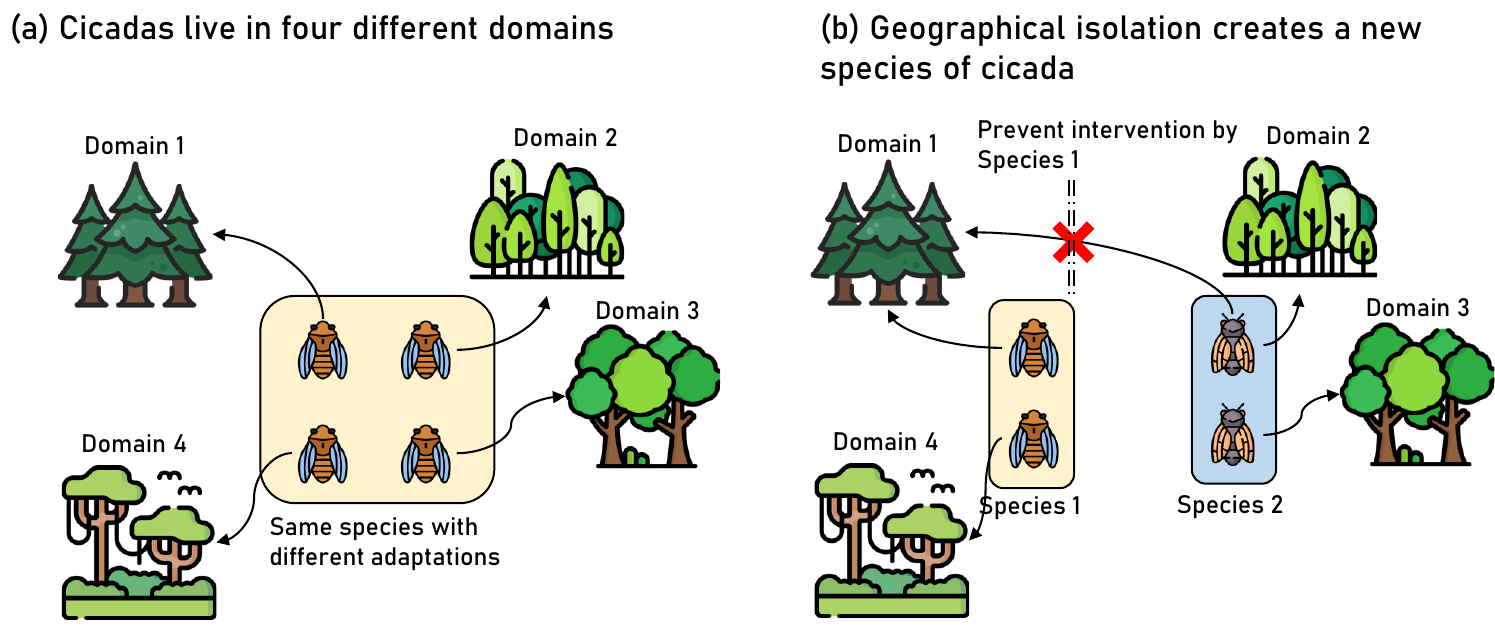}
    \caption{Evolutionary insight for selective meta-learning and adaptive expansion}
    \label{fig:cicada}
    \vspace{-10pt}
\end{figure}

We discuss evolutionary insights to show the rationale behind our selective meta-learning and adaptive expansion. Figures~\ref{fig:cicada}(a) and (b) illustrate the evolution of cicadas, a type of insect. Figure~\ref{fig:cicada}(a) corresponds to Figure~\ref{fig:meta-learning}(c), representing cicadas of the same species(meta-domain) living in different domains with gene flow between them, as reflected by the sharing of the same initial parameters $\Theta_j^1$. However, due to differing environments, they exhibit distinct phenotypes, manifested as different adaptations. Figure~\ref{fig:cicada}(b) corresponds to Figure~\ref{fig:meta-learning}(d), where we artificially introduce geographical isolation for Species 1, resulting in two separate species that adapt to domains 1 and 4, and domains 2 and 3, respectively. There is no gene flow between the two distinct species, which is reflected in the use of different initial parameters. Nevertheless, they have a competitive relationship; specifically, Species 1 prevents Species 2 from invading domains 1 and 4, and Species 2 prevents Species 1 from invading domains 2 and 3, as we show in Section \ref{sec:selective meta-learning}. Finally, the two species follow different evolutionary trajectories because they adapt to different domains.

\end{document}